\documentclass{article}

\usepackage{dilab_arxiv}

\usepackage[capitalize,noabbrev]{cleveref}
\usepackage{enumitem}

\usepackage{graphicx}
\usepackage{url}
\usepackage{booktabs}
\usepackage{multirow}
\usepackage{amsmath,amssymb}
\usepackage{pifont}
\usepackage[table]{xcolor}
\usepackage{makecell}
\usepackage{longtable}
\usepackage{array}
\usepackage{tabularx}
\usepackage{xltabular}
\usepackage{fancyvrb}
\usepackage{listings}

\newcolumntype{Y}{>{\raggedright\arraybackslash}X}

\lstdefinestyle{judgeprompt}{
  basicstyle=\footnotesize\ttfamily,
  breaklines=true,
  breakatwhitespace=false,
  columns=fullflexible,
  keepspaces=true,
  showstringspaces=false,
  frame=none,
  backgroundcolor=\color[gray]{0.95},
  framesep=5pt,
  linewidth=0.94\linewidth,
  xleftmargin=0.03\linewidth,
  xrightmargin=0.03\linewidth,
  aboveskip=0.45em,
  belowskip=0.75em
}

\title{When Personal Memory Has No Single Answer: Evaluating LLM Agents under Irreducible Conflict}
\runningtitle{Evaluating LLM Agents under Irreducible Conflict}
\date{arXiv preprint, \today}

\paperlogo{\includegraphics[height=1.5cm]{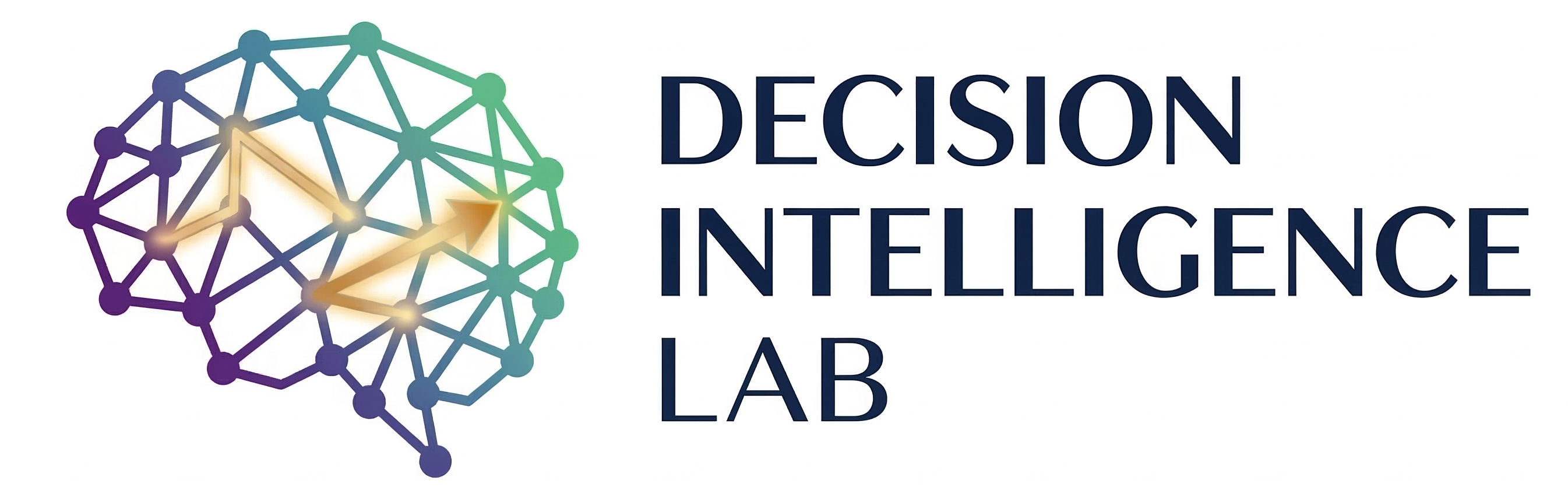}}

\author{
  Lu Yang$^{1}$, 
  Shusheng Xu$^{2}$, 
  Zhuoran Li$^{1}$, 
  Tongkai Yang$^{2}$, and 
  Longbo Huang$^{1}$
  \\[0.3em]\normalfont
  $^1$Institute for Interdisciplinary Information Sciences (IIIS), Tsinghua University \\ $^2$Ant Group
}

\begin{document}

\maketitle
\thispagestyle{fancy}

\begin{abstract}
Large Language Model (LLM)-based agents increasingly maintain persistent personal memory across sessions, but such memory is not always consistent. Preferences depend on context, behaviors evolve over time, and information from different sources can be contradictory. When a query omits the context, time, or source authority needed to interpret conflicting memories, treating one memory as definitive turns an unresolved conflict into an unjustified answer and leads to overconfident actions. Existing benchmarks are built around a single recoverable answer: they reward selecting or reconstructing one value from conflicting evidence, while overlooking whether an agent recognizes underdetermination, preserves alternatives, seeks missing information, and chooses an appropriate action.
We introduce \underline{T}esting \underline{A}gents' \underline{N}avigation of \underline{G}enuine, \underline{L}atent, and \underline{E}ntangled Memory Conflicts (\textsc{TANGLE}), a benchmark for evaluating cognitive behavior when agents face genuinely unresolvable conflicts in personal memory. \textsc{TANGLE} comprises 541 instances that span 40 diverse personas and three conflict types: Context-Partitioned Conflict (CPC), Behavior-Oscillation Conflict (BOC), and Source-Contradiction Conflict (SCC). We evaluate two tracks---an oracle track that provides curated memory directly and a pipeline track in which memory systems extract memory from multi-session dialogues---to assess five cognitive dimensions required for conflict handling: conflict perception, causal reasoning, confidence calibration, clarification seeking, and memory faithfulness.
Our experiments reveal challenges across the agent pipeline. With complete curated memory, models recognize conflicts more reliably than they calibrate actions or seek targeted clarification. With end-to-end pipeline memory, an upstream bottleneck emerges: memory extraction fails to preserve the conflict-bearing relations required for downstream reasoning. At the downstream action stage, policy comparisons show that fixed resolution rules are insufficient when the action must reflect the specific conflict. These findings motivate our Conflict-Aware Action Policy (CAAP), which adapts its action to each conflict according to the available evidence. Together, \textsc{TANGLE} and these findings frame conflict handling as recognizing underdetermination, retaining conflicting evidence, and selecting an action without forcing a definitive answer.
\end{abstract}

\section{Introduction}
\label{sec:intro}

Large language model (LLM) agents are evolving from stateless assistants into persistent systems that retain personal memory across dialogue sessions~\citep{memgpt,mem0,memos}. As memory grows, agents accumulate records about different aspects of a user's life across contexts, over time, and across sources. For a given aspect, conflict arises without any one record being false: a preference is appropriate in one context but not another~\citep{10f513c9-9767-3b38-9731-500af725151f}; a behavior reflects a prior or transient state rather than a stable trait as user information evolves over long-term interaction~\citep{ong-etal-2025-towards,bae-etal-2022-keep,personamem,stale}; and a source is informative within its own scope without being authoritative for every decision~\citep{10.1145/2588555.2610509,TIAN2020104828}.

The challenge is that conflict memory preserves information whose applicability depends on context, time, or source, while the current query omits the variable needed to determine which piece of information applies. The memory bank therefore does not support a unique answer. If an agent nevertheless treats one piece of information as the definitive answer, it removes the conditions under which each piece of information applies, misclassifies the user's current state, gives advice that is inappropriate for the present situation, or executes a choice that has not been confirmed. This makes conflict handling a central reliability and safety problem for memory agents~\citep{persistbench,memevobench}: consequential settings require cautious verification, while lower-stakes settings call for conditional or reversible assistance.

Existing conflict benchmarks primarily focus on recovering a unique answer when the available evidence supports one, leaving behavior under genuinely unresolvable memory conflict outside their evaluation target. Their task definitions formulate conflict as a problem with a single recoverable answer, treating disagreement as something to settle rather than structure to preserve. This answer-centered design encourages agents to force conflicting memories into one definitive answer, producing overconfident or inappropriate actions when the relevant context, temporal state, or source authority remains unresolved. In addition, prior benchmarks do not disentangle memory extraction from conflict resolution. A single, undifferentiated track conflates failures in memory construction with failures in conflict reasoning, leaving unclear whether an agent lacks conflict-bearing evidence or fails to act appropriately on the evidence it receives. The comparison in Table~\ref{tab:benchmark_comparison} makes the resulting gap concrete: prior benchmarks do not jointly support multiple valid responses, qualitative response grading, multiple evaluation tracks, and genuine ambiguity.

\providecommand{\yes}{{\color{green!60!black}\ding{51}}}
\providecommand{\no}{{\color{red!65!black}\ding{55}}}

\begin{table*}[h]
  \centering
  \caption{Comparison with prior conflict benchmarks. Ctx: average/maximum context length; \#Q: queries; MV/QL/MS/MT/Amb.: multiple valid responses, graded quality, multi-session data, multi-track inputs, and genuine ambiguity.}
  \label{tab:benchmark_comparison}
  \scriptsize
  \setlength{\tabcolsep}{3.5pt}
  \renewcommand{\arraystretch}{1.12}
  \begin{tabularx}{\textwidth}{@{}l c r >{\raggedright\arraybackslash}X >{\raggedright\arraybackslash}X c c c c c@{}}
    \toprule
    \multirow{2}{*}{\textbf{Benchmark}} &
    \multirow{2}{*}{\makecell{\textbf{Ctx}\\\textbf{(avg./max.)}}} &
    \multirow{2}{*}{\textbf{\#Q}} &
    \multirow{2}{*}{\makecell[l]{\textbf{Conflict}\\\textbf{focus}}} &
    \multirow{2}{*}{\makecell[l]{\textbf{Response}\\\textbf{format}}} &
    \multicolumn{5}{c}{\textbf{Evaluation coverage}} \\
    \cmidrule(l){6-10}
    & & & & & \textbf{MV} & \textbf{QL} & \textbf{MS} & \textbf{MT} & \textbf{Amb.} \\
    \midrule
    MemConflict~\citep{MemConflict} & 3.9K/204K & 1,492 & Temporality / fact / context & Short answer & \no & \no & \yes & \no & \no \\
    STALE~\citep{stale} & 152K/165K & 1,200 & Implicit temporality & Open-form & \no & \no & \yes & \no & \no \\
    HaluMem~\citep{halumem} & 2.3K/1M & 3,467 & Erroneous premise & Short answer & \no & \no & \yes & \no & \no \\
    ConflictBank~\citep{conflictbank} & N/A$^\dagger$ & 553K & Fact & Multiple choice & \no & \no & \no & \no & \no \\
    CONFLICTINGQA~\citep{conflictingqa} & 512/1K & 238 & Opinion & Binary choice & \no & \no & \no & \no & \no \\
    CONFLICTS~\citep{draggedconflicts} & 512/5K & 458 & Temporality / fact / opinion & Open-form & \no & \no & \no & \no & \no \\
    SelectiveQA~\citep{selectiveqa} & N/A$^\ddagger$ & 34.6K & Source bias & Short answer + abstain & \no & \no & \no & \no & \no \\
    \rowcolor[gray]{0.93}
    \textbf{TANGLE} & \textbf{744/48K} & \textbf{541} & \textbf{Context / behavior / source} & \textbf{Open-form} & \yes & \yes & \yes & \yes & \yes \\
    \bottomrule
  \end{tabularx}

  \vspace{2pt}
  \parbox{\textwidth}{\scriptsize $^\dagger$ Sentence-level claim--evidence pairs. \quad $^\ddagger$ Structured tabular input.}
\end{table*}

To address this gap, we introduce \underline{T}esting \underline{A}gents' \underline{N}avigation of \underline{G}enuine, \underline{L}atent, and \underline{E}ntangled Memory Conflicts (\textsc{TANGLE}), a benchmark that evaluates how an agent shall reason and act when personal memory contains a genuinely unresolvable conflict. \textsc{TANGLE} comprises 541 instances from 40 personas, covering 46 life aspects in 10 domains and three conflict types: \emph{Context-Partitioned Conflict (CPC)}, in which a preference depends on context; \emph{Behavior-Oscillation Conflict (BOC)}, in which preferences change without stabilizing; and \emph{Source-Contradiction Conflict (SCC)}, in which sources disagree. These conflicts are unresolvable because context, behavioral dynamics, or source reliability is unknown. We evaluate \textsc{TANGLE} through two matched tracks. The \emph{oracle} track provides curated memory directly and isolates conflict reasoning. The \emph{pipeline} track evaluates end-to-end behavior after a memory system extracts memory from multi-session dialogue~\citep{locomo,longmemeval}. Because both tracks encode the same underlying conflicts, their comparison separates information loss during memory construction from failures in conflict reasoning. Since no response is uniquely correct, our rubric assesses conflict perception, causal reasoning, confidence calibration~\citep{uncertaintydecomposition}, clarification seeking, and memory faithfulness, and flags high-risk cases where overconfident resolution causes harm. Representative instances are provided in Appendix~\ref{app:examples}.

Across five response models, oracle results reveal a recognition-to-action gap: even with complete curated memory, models recognize conflicts more reliably than they calibrate recommendations or seek targeted clarification. Comparing the oracle and pipeline tracks identifies memory extraction as an upstream bottleneck, as systems often retain topical facts while losing the context, temporal, and source relations required to interpret the conflict; partial retrieval consequently lowers perception and diagnosis. The loss of conflict structure, together with same-domain distractors, further weakens subsequent reasoning and action. Policy comparisons show that deterministic selectors and conservative templates impose predetermined response patterns on unresolved conflicts. These findings motivate Conflict-Aware Action Policy (CAAP), which adapts its action to each conflict according to the available evidence.

Our contributions are:
\begin{itemize}
    \item We expose two limitations in prior memory-conflict benchmarks: they formulate conflict as a single-answer recovery task and evaluate agents in a single, undifferentiated track. The single-answer formulation encourages agents to force conflicting memories into one definitive answer, producing overconfident or inappropriate actions when the relevant context, temporal state, or source authority is unresolved. The single-track design conflates failures in memory construction with failures in conflict reasoning, leaving unclear whether an agent lacks conflict-bearing evidence or fails to act appropriately on the evidence it receives.
    \item We establish \textsc{TANGLE}, a benchmark that shifts personal-memory conflict evaluation from recovering a single answer to assessing how agents recognize underdetermination, retain conflicting evidence, and choose an appropriate action. It contains 541 instances across 40 personas, 46 life aspects, 10 domains, and three conflict types. We provide oracle and pipeline tracks that separate conflict reasoning from memory extraction, together with a behavior-centered rubric for cases without a unique gold answer.
    \item We show that fixed resolution rules provide an incomplete treatment of unresolved conflict, and introduce Conflict-Aware Action Policy (CAAP), which selects an action for each conflict from the visible evidence rather than applying a single resolution rule.
\end{itemize}

\section{Related Work}
\label{sec:related}

\subsection{Conflict Benchmarks}

Work on conflict evaluation spans settings with deterministic resolution and settings that characterize model behavior under disagreement. Many benchmarks formulate conflict resolution as recovering a single answer, whether through temporal recency~\citep{MemConflict,membench}, implicit state invalidation~\citep{stale}, premise correction~\citep{halumem}, explicit resolution mandates~\citep{memoryagentbench}, pre-construction conflict elimination~\citep{mem2actbench}, or knowledge-grounded contradiction handling~\citep{wikicontradict,conflictqa}. A parallel line instead examines how models arbitrate among competing evidence, including memorization tendencies under forced choice~\citep{conflictbank,adaptivechameleon}, source and evidence sensitivity~\citep{conflictingqa,whosfactswin}, dynamic fact updates and misinformation effects~\citep{DYNAMICQA,misbench}, conflict-type-aware response generation~\citep{draggedconflicts}, multi-hop conflict localization~\citep{MAGIC}, and selective question answering (QA) with abstention over conflicting multi-source personal memory~\citep{selectiveqa}. For broader context, \citet{knowledgeconflictssurvey} provide a comprehensive synthesis. Our evaluation focuses specifically on the quality of model behavior when conflicts are intentionally unresolvable by design.

\subsection{Long-Term Memory Benchmarks}

Long-term memory benchmarks have largely emphasized stable recall, personalization fidelity, and memory use under evolving interaction histories. Prior datasets test factual and temporal recall~\citep{locomo,longmemeval}, preference consistency and persona adherence~\citep{prefeval,personamem,PersonaMem-v2}, and extend to real-dialogue personalization~\citep{AlpsBench} and emotionally supportive conversations~\citep{es-memeval}. Recent work increasingly stresses dynamic memory conditions, including forgetting-aware evaluation~\citep{memora}, memory-to-action grounding~\citep{mem2actbench}, multi-session state tracking~\citep{memoryarena,memtrack}, and self-evolving memory~\citep{evomembench,evomemory,memevobench}, while \citet{persistbench} highlights downstream safety concerns such as memory-induced sycophancy.

\subsection{Agent Memory Systems}

Agent memory architectures span virtual context management and scalable external memory layers~\citep{memgpt,mem0,memos,memorybank}, reflective and self-updating memory mechanisms~\citep{amem,meminsight}, and procedural memory for workflow execution~\citep{memp,agentworkflowmemory}. Optimization-oriented approaches further treat memory use as a policy-learning problem, including reinforcement-based methods for memory control~\citep{memoryr1,memrl,memento}.


\begin{figure*}[t]
    \centering
    \includegraphics[width=\textwidth]{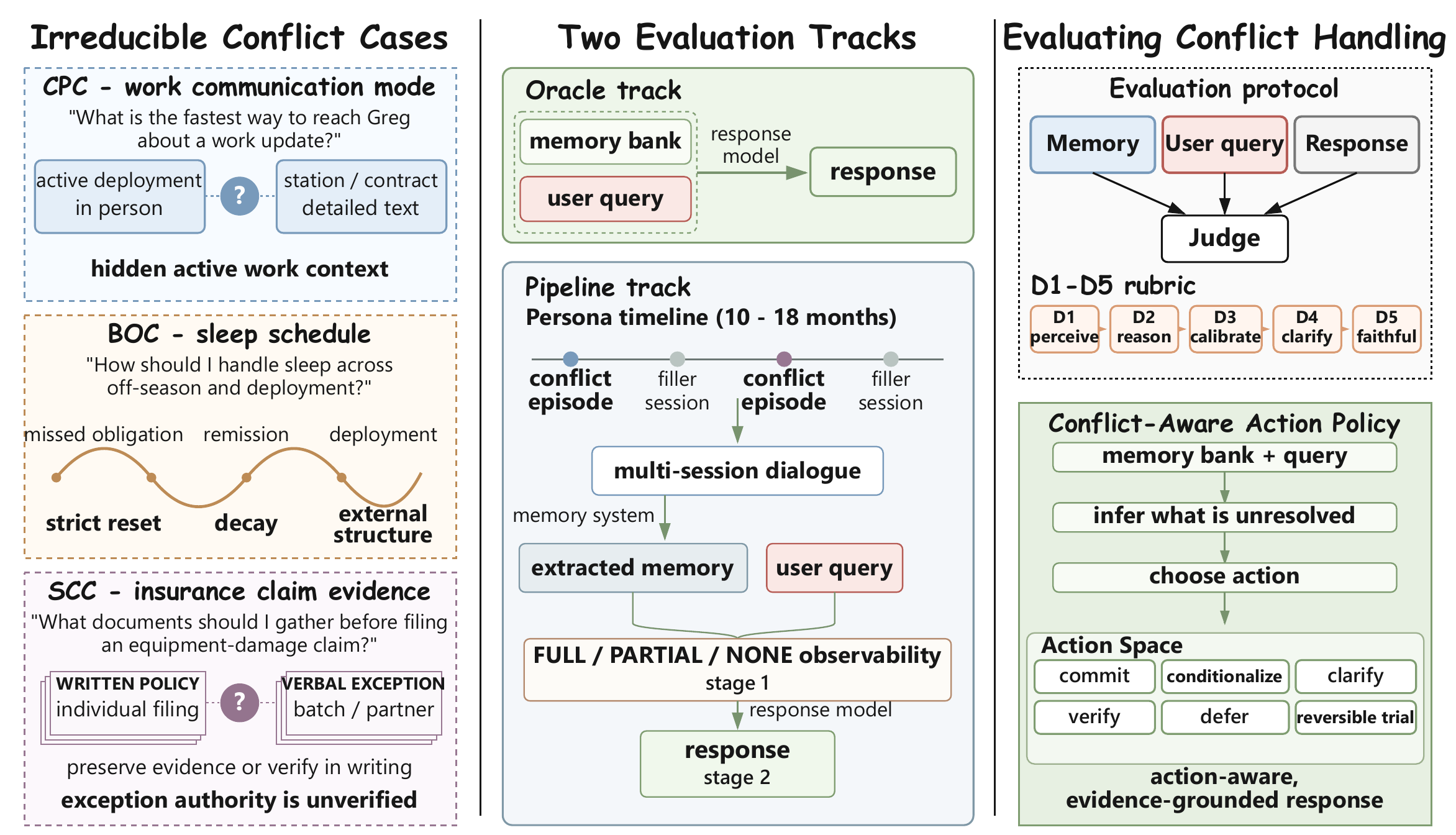}
    \caption{\textbf{Overview of the \textsc{TANGLE} benchmark.} \textsc{TANGLE} (\underline{T}esting \underline{A}gents' \underline{N}avigation of \underline{G}enuine, \underline{L}atent, and \underline{E}ntangled Memory Conflicts) covers Context-Partitioned Conflict (CPC), Behavior-Oscillation Conflict (BOC), and Source-Contradiction Conflict (SCC), with oracle and pipeline tracks evaluating agents' capabilities in handling memory conflicts. The Conflict-Aware Action Policy (CAAP) selects evidence-grounded actions rather than forcing a single value.}
    \label{fig:overview}
\end{figure*}
\begin{figure*}[t]
    \centering
    \includegraphics[width=\textwidth]{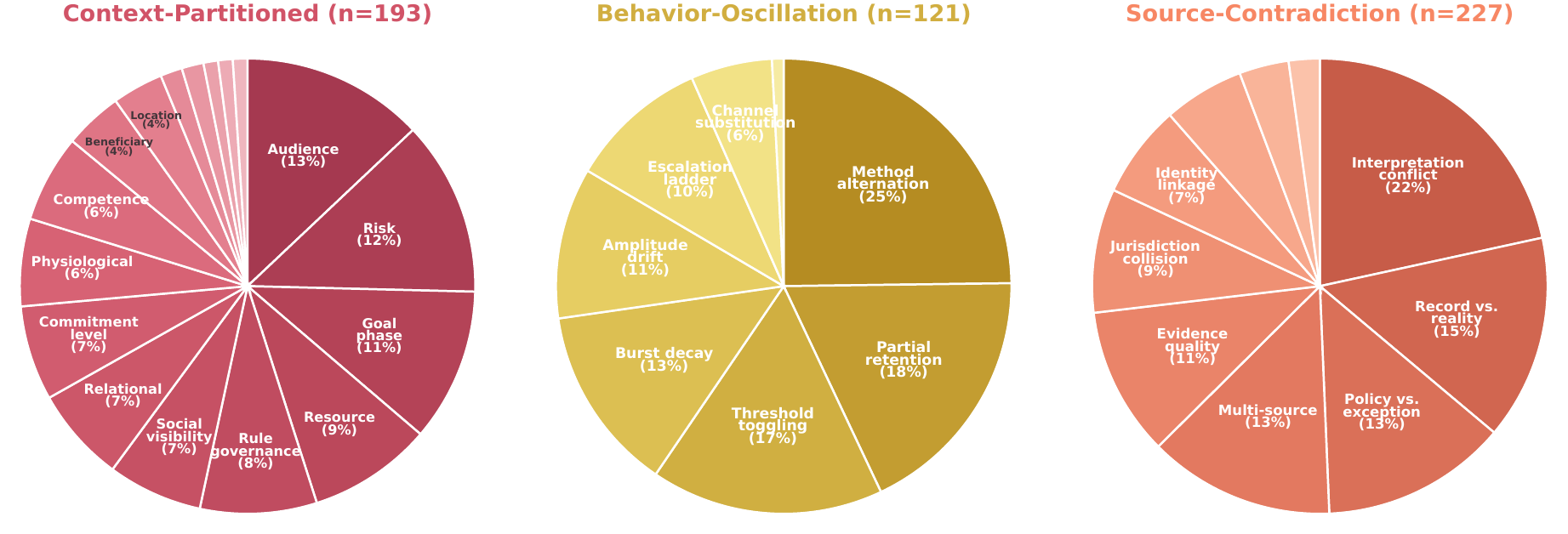}
    \caption{\textsc{TANGLE} structural diversity by conflict type. The three panels show the distributions of CPC context-partition types, BOC oscillation shapes, and SCC source-conflict types; each labeled sector reports its share within that conflict type.}
    \label{fig:benchmark_taxonomy}
\end{figure*}

\section{The \textsc{TANGLE} Benchmark}
\label{sec:benchmark}

\subsection{Problem Formulation}

The goal of our benchmark is to evaluate how effectively LLM agents resolve conflicts in personal memory where the available evidence does not support a unique deterministic answer. Let $\mathcal{M} = \{m_1, \dots, m_n\}$ denote a memory bank comprising conflict-bearing memories and background entries, and let $\mathcal{D} = \{d_1, \dots, d_k\}$ denote same-domain distractor memories that are topically related but non-diagnostic for the conflict. Given the combined input $(\mathcal{M} \cup \mathcal{D}, q)$---where $q$ is a natural user request that does not indicate which information is currently relevant---the agent produces a response $r$.

In deployed memory agents, conflict resolution failures stem from two distinct sources: the memory system fails to extract or retrieve relevant evidence, or the model fails to reason appropriately over the evidence it receives. We therefore define two evaluation tracks. Under the \emph{oracle track}, $r = f(\mathcal{M} \cup \mathcal{D},\; q)$: the model directly receives the curated memory set and must resolve the conflict from the evidence alone. Under the \emph{pipeline track}, a memory system $g$ first processes multi-session dialogue $S = (s_1, \dots, s_T)$ to produce an extracted memory set $\hat{\mathcal{M}} = g(S)$; the model then responds as $r = f(\hat{\mathcal{M}},\; q)$. Comparing oracle and pipeline performance isolates conflict resolution capability from extraction quality.

Unlike prior benchmarks where a unique gold output $y^*$ exists, our conflicts are constructed so that multiple response strategies can all be appropriate. Effective resolution demands the following cognitive capabilities:
\textbf{1) Conflict awareness.} Recognizing that the memory bank contains contradictory or incompatible information relevant to the query.
\textbf{2) Causal reasoning.} Analyzing why the conflict exists---comparing hypotheses and linking reasoning to specific memory evidence.
\textbf{3) Epistemic calibration.} Expressing certainty proportional to evidence strength---committing when evidence is clear, hedging when genuinely ambiguous.
\textbf{4) Information seeking.} Identifying what missing information would resolve the conflict and asking targeted questions.
\textbf{5) Evidential grounding.} Basing all factual claims on the memory bank without fabricating information beyond what is stored.

\subsection{Data Statistics and Conflict Taxonomy}

The benchmark contains 541 instances spanning 40 personas, 46 life aspects (recurring facets of a user's life---behaviors, preferences, and factual records such as transport mode, exercise routine, and financial records) across 10 life domains and three conflict types (Figure~\ref{fig:benchmark_taxonomy}): Context-Partitioned Conflict (CPC, 193 instances), Behavior-Oscillation Conflict (BOC, 121 instances), and Source-Contradiction Conflict (SCC, 227 instances). Under the oracle track, each instance contains 8--11 core memory items plus 6 same-domain distractor memories and a natural user query, totaling 4,893 core memories and 3,246 distractors across the benchmark. Under the pipeline track, we provide 2,580 multi-session dialogues (mean 64.5 sessions per persona, mean 744 tokens per session, $\sim$48K tokens per persona) from which memory systems extract user information before receiving the same queries.

Each conflict type is defined by a distinct latent variable that makes deterministic resolution impossible. \textbf{(1) Context-Partitioned Conflict (CPC).} Let $\mathcal{M}=\{m_i\}$ be the set of retrieved memories, where each memory $m_i$ states a value $v_i$ for the same target life aspect $a$ (e.g., preferred restaurant type). In CPC, each statement is valid only under an unobserved context $c$ (such as audience or risk level): $m_i \text{ holds} \iff c \in C_i$, where the context regions $\{C_i\}$ are disjoint and support incompatible values for $a$. If the query $q$ does not specify which context applies, no single value is entailed. Our CPC subset covers 17 context-partition types (including audience, goal phase, risk level, and social visibility). \textbf{(2) Behavior-Oscillation Conflict (BOC).} Memories in $\mathcal{M}$ report the user's value on life aspect $a$ at different times, denoted $v_{t_1}, v_{t_2}, \dots$ (the observed value of $a$ at successive time points $t_1, t_2, \dots$), but the trajectory is non-convergent---exhibiting repeated reversals, partial retention, or amplitude drift rather than settling to a stable state. The latent factor is the user's oscillation phase $z_t \in \{$adopt, friction, abandon, retry$, \dots\}$; when $q$ does not anchor the relevant phase, the conflict is unresolvable from any single memory. Our BOC subset includes 13 oscillation drivers and 8 oscillation shapes. \textbf{(3) Source-Contradiction Conflict (SCC).} Memories report conflicting values for the same life aspect $a$ from $K \ge 2$ different sources $s_1, \dots, s_K$ (e.g., self-report, behavioral inference, official record): each memory $m_j$ reports a value $v_j$ from source $s_j$, and these values disagree ($v_j \neq v_k$ for $j \neq k$). The latent variable is source reliability $w(s)$; if $q$ does not indicate which source governs and the reliability ordering is itself uncertain, no fixed priority rule yields a deterministic answer. Our SCC subset contains 10 source-conflict types, with source-count distribution of 2 sources (54\%), 3 sources (37\%), and 4 sources (10\%).

\subsection{Data Construction}

\textbf{Persona--aspect schema.} Starting from Persona Hub~\citep{personahub} seeds, we sample and curate a diverse pool of 40 personas spanning age, occupation, income, family structure, health conditions, and geographic setting. Persona Hub supplies initial profile material rather than fixed benchmark instances; we normalize the sampled profiles into a common schema and use internal benchmark identifiers because no formal source IDs are retained. Each persona is mapped to a sparse subset of relevant life aspects rather than a full Cartesian assignment, preserving realism under life constraints (e.g., a rotating-shift nurse receives \emph{sleep\_schedule} but not \emph{career\_commitment}). Full schema definitions are in Appendix~\ref{app:personas}--\ref{app:attributes}.

\textbf{Instance synthesis.} For each selected persona--aspect pair, we generate conflict instances with same-domain distractors using a constrained generator. Memories are written as third-person episodic statements compressed to 18--32 words, matching the output format of deployed memory systems. We enforce two integrity constraints: (i) a \emph{non-leakage} rule guarantees no single memory states the conclusion---the conflict is only recoverable from the aggregate evidence; (ii) \emph{pattern balancing} ensures no single conflict subtype dominates through targeted diversity rewrites. Generation details and quality control procedures are in Appendix~\ref{app:generation_qc}.

\textbf{Query protocol.} Each instance is paired with a natural user request that reads as an ordinary task yet withholds the information needed to resolve the conflict. CPC queries hide the context variable entirely, forcing the agent to ask or conditionalize. BOC and SCC queries use open-ended task delegation (e.g., ``help me get my exercise routine working well'') without temporal anchors or explicit conflict cues, requiring proactive conflict discovery.

\textbf{Multi-session dialogue synthesis (pipeline track).} The pipeline track is derived from the same curated oracle memory set: for each instance we verbalize its memories into longitudinal multi-session dialogue, rendering every memory as user utterances within natural multi-turn conversations that stay consistent with the persona's profile and speaking style. Conflict-bearing memories are staggered across sessions to avoid clustering, and filler sessions containing no conflict-relevant information are interleaved. Because both tracks encode identical underlying conflicts, comparing them isolates extraction quality from conflict resolution. Full generation algorithm and prompts are in Appendix~\ref{app:prompts}.

\subsection{Rubric Design}

Because our conflicts admit no single gold answer, we score the \emph{quality of an agent's cognitive behavior} rather than the correctness of a final value. We operationalize the five capabilities required for effective resolution into five scoring dimensions (D1--D5), each rated on a $0$--$4$ scale with conflict-type-specific anchors. Dimensions are scored independently, so a response may score high on one and low on another. \textbf{D1 (Conflict Perception)} captures whether the response identifies and articulates the conflict. \textbf{D2 (Causal Reasoning)} captures the depth of causal or source-reliability reasoning linked to specific memory evidence rather than generic explanation. \textbf{D3 (Confidence Calibration)} captures whether expressed certainty matches evidence strength, rewarding conditional recommendations over overcommitment. \textbf{D4 (Clarification Seeking)} captures the quality of information-seeking targeted at the unresolved conflict variable. \textbf{D5 (Memory Faithfulness)} captures whether factual claims are traceable to the memory bank without fabrication. For the 62 high-risk SCC instances in medical and authorization domains, we additionally assign a ternary \textbf{D6 (Commitment Appropriateness)} flag---\textsc{safe}, \textsc{partial}, or \textsc{unsafe}---reported separately from the composite score. Full dimension definitions with anchoring examples are in Appendix~\ref{app:rubric}.

\section{Experiments}
\label{sec:experiments}

We use \textsc{TANGLE} to study how well current LLM agents handle genuinely unresolvable conflicts in personal memory. Rather than asking whether a model recovers a single correct value, we measure the quality of the cognitive behavior it exhibits: whether it perceives the conflict, reasons about its source, calibrates its confidence, seeks the information needed to resolve the conflict, and stays faithful to the memory bank. Our experiments are organized around four questions that move from model
capability to system reliability and policy design:

\begin{itemize}[leftmargin=1.5em,itemsep=2pt,topsep=2pt]
\item What profiles do current LLM agents exhibit when handling genuinely unresolved conflicts, and how do these profiles vary across conflict types, cognitive dimensions, and models? (\S\ref{sec:rq1})
\item How does memory construction shape the availability and quality of conflict-sensitive reasoning in an end-to-end system? (\S\ref{sec:rq2})
\item How does conflict handling change as memory noise increases? (\S\ref{sec:rq3})
\item What do different conflict-handling policies reveal about resolving unresolved conflict? (\S\ref{sec:rq4})
\end{itemize}

\subsection{Setup}

We evaluate Claude Sonnet 5, Gemini 3.1 Pro, DeepSeek-V3.2, GLM-4.7, and GPT-4o in the \emph{oracle} and \emph{pipeline} tracks. Unless otherwise stated, models generate responses with temperature 0 and a 4,096-token limit. GPT-5.4 and Claude Opus 4.7 independently score each response on D1--D5 using the same 0--4 rubric, and we report their results separately throughout; their total scores agree strongly across the 2,705 shared responses (Pearson $r=0.838$; Spearman $\rho=0.820$). To assess rubric reliability, two human annotators independently score a 556-record reference set and agree within one point on 85.8\% of dimension scores overall, 89.4\% for BOC, 80.5\% for SCC, and 91.3\% for CPC.

\subsection{Response Quality in the Oracle Track}
\label{sec:rq1}

\begin{figure*}[t]
    \centering
    \includegraphics[width=0.98\textwidth]{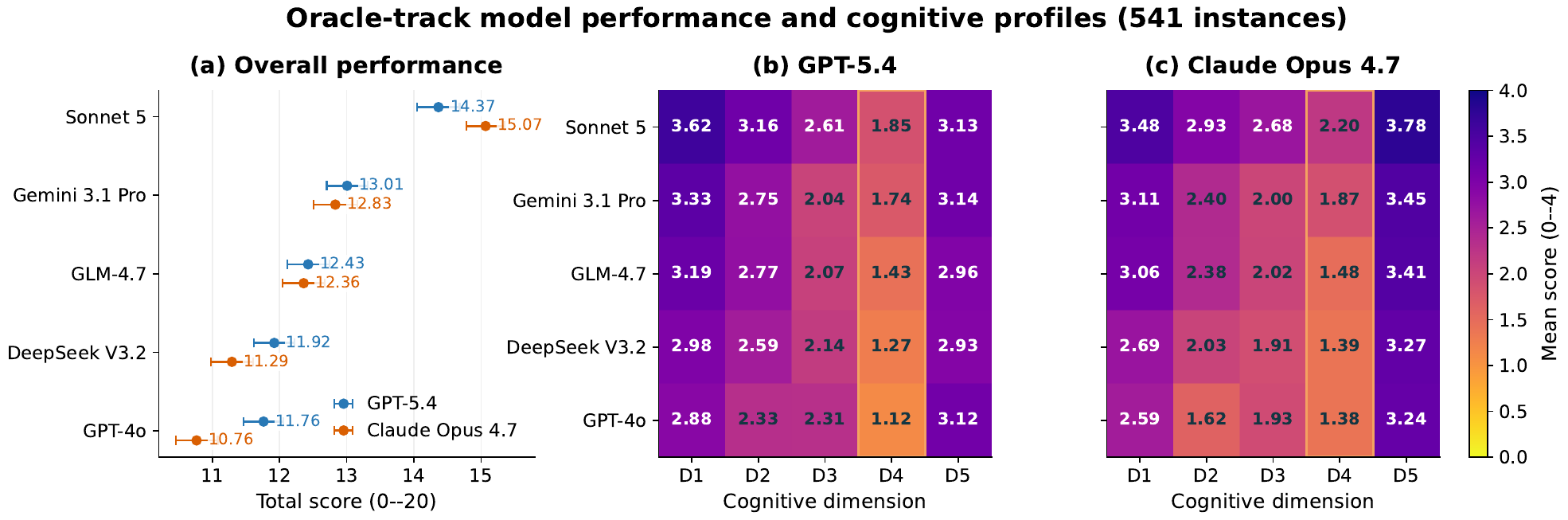}
    \caption{\textbf{Oracle-track performance and cognitive profiles.} (a) Mean D1--D5 totals with 95\% confidence intervals. (b,c) Judge-specific model--dimension profiles (0--4); D4 clarification is the main bottleneck.}
    \label{fig:rq1_oracle_summary}
\end{figure*}

We first evaluate response quality when every model receives the complete oracle memory bank: all canonical core memories and six same-domain distractors for each of the 541 instances. The aggregate totals indicate substantial room for improvement even under complete oracle access: the strongest model remains below 15/20 under both Judges. Figure~\ref{fig:rq1_oracle_summary} reports the overall model ranking and the model--dimension profiles under both Judges; no separate oracle table is needed.

\textbf{Conflict structure determines difficulty.} Performance varies systematically with the unresolved variable: CPC is easiest (14.79 under GPT-5.4; 14.96 under Opus), SCC is intermediate (11.92; 11.58), and BOC is hardest (10.81; 10.14). CPC requires reasoning about which context--preference mapping applies when the query omits the relevant context. SCC requires comparing conflicting claims whose source scope, authority, or evidence quality differ. BOC requires reasoning about temporal stability rather than selecting among competing propositions: the model must determine whether the current behavioral state will persist. The conflict-type comparison is summarized numerically above, while Figure~\ref{fig:rq1_oracle_summary}(b,c) shows the corresponding model--dimension profiles under the two Judges.

\textbf{Models recognize conflict more reliably than they act under it.} Across both Judges, conflict perception, diagnosis, and memory faithfulness are stronger than confidence calibration and clarification seeking, with D4 the lowest-scoring dimension for every response model. The common bottleneck is therefore not simply detecting contradictory evidence or avoiding fabrication. Models often describe the tension without expressing calibrated uncertainty, asking the question that would change the recommendation, or selecting a reversible action appropriate to the unresolved conflict.

BOC provides the clearest evidence of this gap: D5 remains high (3.52--3.55), while D3 and D4 remain low (D3: 0.65--1.03; D4: 0.65--0.69). Models faithfully recount an oscillatory history without treating that history as evidence that a fixed recommendation does not persist. Detecting behavioral change does not by itself tell the model whether the current state will persist, whether a proposed intervention will work, or which unresolved trigger it needs to clarify. Figure~\ref{fig:rq1_oracle_summary} summarizes the model-level dimension profile.

\textbf{Model profiles differ within this shared bottleneck.} Although all models exhibit the recognition-to-action gap, its severity varies. Sonnet~5 is the most balanced model across dimensions, whereas Gemini~3.1~Pro and GLM-4.7 retain relatively strong conflict perception but are less consistent in translating that recognition into calibrated action. DeepSeek~V3.2 and GPT-4o show the weakest clarification behavior, with GPT-4o also exhibiting a diagnosis deficit on source-contradiction cases. Thus, the models share the same broad recognition-to-action gap, but differ in which stage of conflict resolution is most affected.

\subsection{Oracle vs. Pipeline Degradation}
\label{sec:rq2}

The oracle track supplies the complete curated memory bank, whereas the pipeline track uses the ordered memories returned by a system-native extractor. We first quantify how often each system preserves the target conflict using three observability states: \emph{FULL}, when the complete conflict structure is visible; \emph{PARTIAL}, when target-specific tension remains visible but at least one conflict-bearing relation is missing or degraded; and \emph{NONE}, when the retrieved bank does not expose the target conflict, even if it contains useful memories for answering the query. We then evaluate downstream response quality with D1--D2 conditioned on the
observability state and D3--D5 computed over all retrieved banks. This keeps
conflict perception and diagnosis separate from calibration, clarification,
and faithfulness, which remain applicable even when the target conflict is
not fully exposed.

\begin{figure*}[t]
    \centering
    \includegraphics[width=0.96\textwidth]{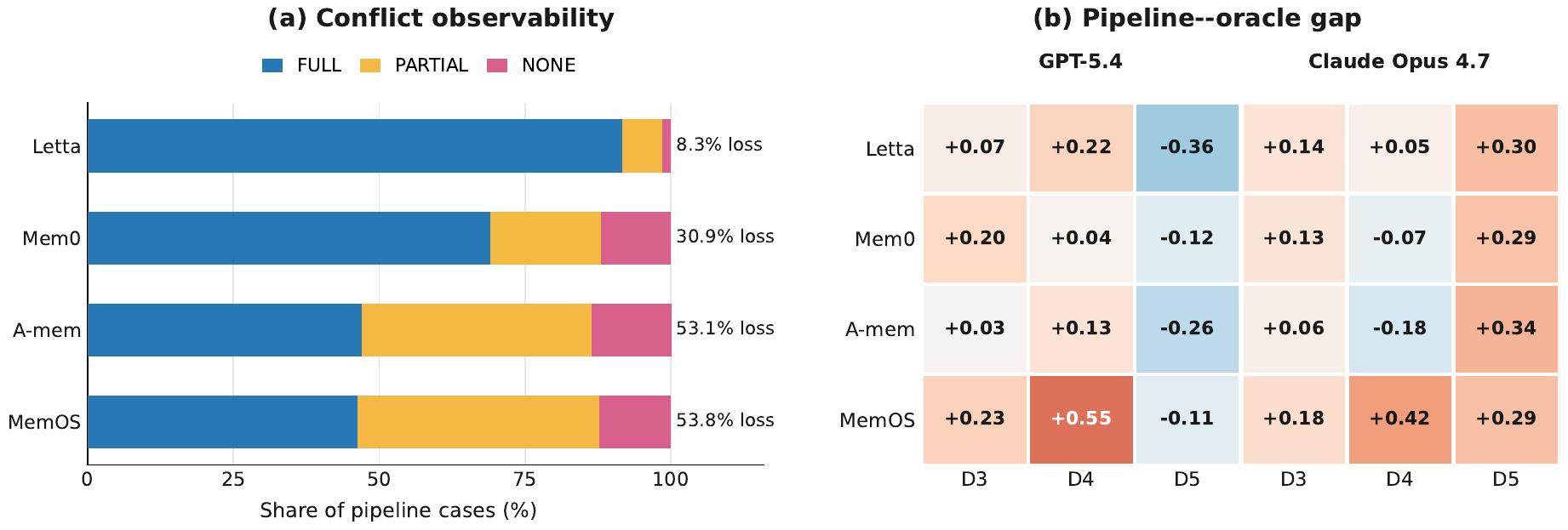}
    \caption{\textbf{Memory construction shapes observability and action.} (a) Pipeline extraction yields full, partial, or no target-conflict observability. (b) Conditional pipeline-minus-oracle gaps for D3--D5; negative values favor oracle inputs.}
    \label{fig:rq2_summary}
\end{figure*}

\textbf{Memory systems differ sharply in conflict observability.} Letta preserves the complete target conflict in 91.7\% of cases, compared with 69.1\% for Mem0, 47.0\% for A-mem, and 46.2\% for MemOS (Figure~\ref{fig:rq2_summary}(a)). The resulting information loss ranges from 8.3\% to 53.8\%. This is a retrieval-level result rather than a downstream quality ranking: a non-FULL bank still supports a useful answer while omitting the trajectory, context partition, or source comparison that defines the benchmark conflict.

\textbf{Complete and partial retrieval reveal different downstream limits.} Figure~\ref{fig:rq2_summary}(b) shows that
pipeline responses do not simply incur a uniform penalty relative to the
complete-evidence oracle. Across systems, D5 remains comparatively robust,
whereas D3 and especially D4 are more variable; this indicates that models
remain faithful to the memories they receive while still lacking the
calibration or targeted inquiry needed to resolve an incompletely observed
conflict. The oracle therefore provides an upper-condition reference for
response behavior, not a single expected score that every pipeline system matchs.

Conditional on FULL retrieval, D1 and D2 are relatively similar across memory
systems under both Judges; the range across systems is much smaller than the
corresponding gap in FULL rates. Once the target conflict has been
reconstructed, the response models therefore receive broadly comparable
evidence for perception and diagnosis. PARTIAL retrieval provides a useful
boundary condition: although D1 and D2 are scored only on the tension that
remains visible, both Judges report lower values than under FULL retrieval for
every system. The lower PARTIAL values are explained by a structural loss: the model must
interpret the surviving memories without the relation that links them to the
omitted evidence, making both conflict mapping and diagnosis less determinate.

The axis-level breakdown makes the consequence concrete. BOC requires a
multi-phase behavioral trajectory, CPC requires both sides of a context
partition, and SCC requires an attributed map of competing claims and their
sources. Partial retrieval preserves answer-relevant content while
destroying the relation that determines how that content is
interpreted. If an SCC bank retains one claim but omits the competing source,
for example, the model produces a locally coherent recommendation based
on the surviving claim but cannot compare the claims or determine whether
their reliability or scope differs. Similarly, omitting earlier BOC phases
hides the trajectory that makes the current state uncertain, while omitting
one CPC context hides the condition under which the alternative preference
applies. The consistency of this pattern across response models indicates that the
missing relations change the reasoning problem itself: each model must
interpret a fragment whose meaning is underdetermined by the retrieved bank.

\textbf{Extraction is the main pipeline bottleneck.} The system-level gaps are driven primarily by whether the memory bank
preserves the relations that define the target conflict. Letta reaches the
FULL condition far more often than the other systems, whereas the
conditional D1/D2 differences among FULL cases are comparatively modest.
Thus, improving end-to-end conflict handling requires more than retrieving
topically relevant memories. It requires preserving the conflict-bearing
relations, dependencies, and evidence structure needed for the response model
to reconstruct, explain, and act on unresolved tension. The response model is unchanged across systems, but the evidence available to
it is not. The dominant failure point therefore lies in memory construction
and conflict observability.

\begin{figure*}[t]
    \centering
    \includegraphics[width=0.96\textwidth]{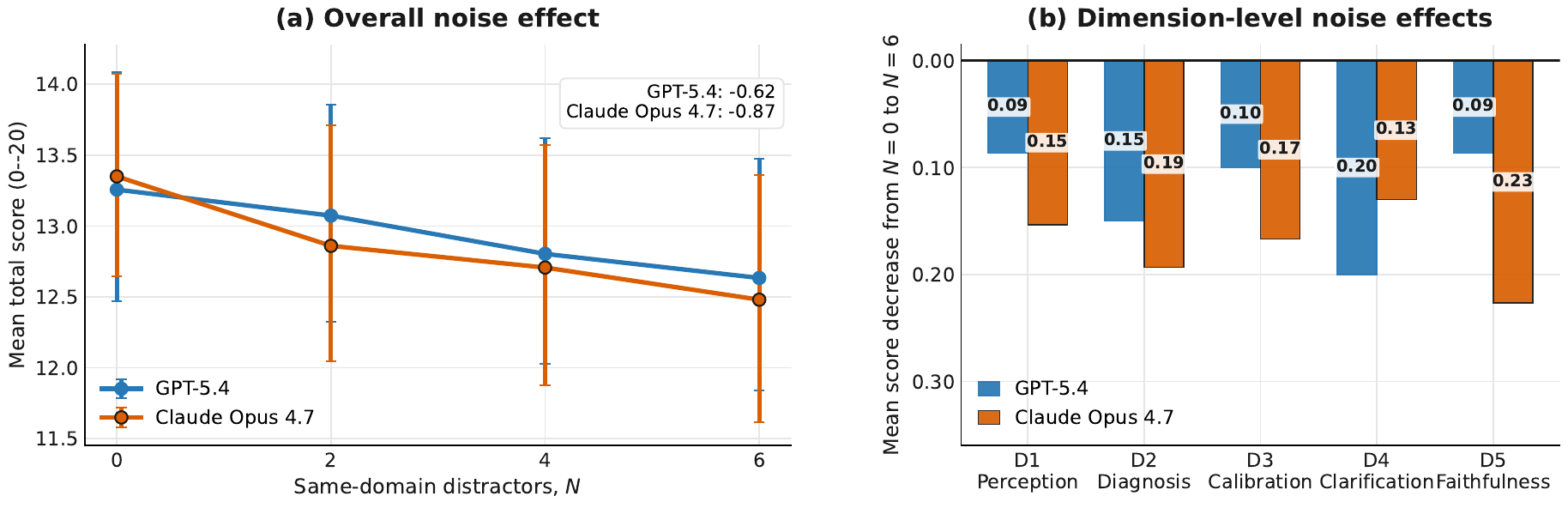}
    \caption{\textbf{Distractor noise weakens conflict-sensitive reasoning.} (a) Mean total quality from $N=0$ to $6$ distractors, with 95\% intervals. (b) Dimension-wise score drops; larger values indicate greater degradation.}
    \label{fig:rq3_summary}
\end{figure*}
\subsection{Conflict-Sensitive Reasoning under Distractor Memories}
\label{sec:rq3}

We next compare response quality across distractor-load conditions using a paired cohort of 60 instances, balanced across BOC, CPC, and SCC. For each instance, we retain the conflict-bearing memories and vary the number of same-domain distractors included in the response-model input, with $N\in\{0,2,4,6\}$. The $N=6$ condition is the standard oracle input, while $N<6$ conditions are lower-load ablations. Each condition contains five response models, and all comparisons are paired within instance. We report Judge-specific estimates and use paired instance-cluster bootstrap intervals to quantify uncertainty in the within-instance changes.

\textbf{Distractors impose a cumulative cost on conflict handling.} The aggregate D1--D5 score total declines monotonically with distractor load under both evaluations: the mean total decreases from 13.26 to 12.63 under GPT-5.4 and from 13.35 to 12.48 under Opus when moving from $N=0$ to $N=6$. The corresponding paired intervals exclude zero for both evaluations. We therefore examine the dimension-level profiles separately to distinguish degradation in conflict-sensitive reasoning from changes in faithfulness to the memory bank. Figure~\ref{fig:rq3_summary}(a) shows a gradual decline rather than a sharp threshold at one particular load. Increasing the memory set therefore creates a competition problem: topically related memories remain plausible user evidence, but make it harder to preserve the distinction between information that defines the conflict and information that only describes the broader user profile.

\textbf{Noise weakens post-detection reasoning more consistently than initial conflict perception.} We separate the conflict-sensitive dimensions (D1--D4) from faithfulness to the visible memory bank (D5). D1 changes only modestly as distractors increase, whereas D2--D4 show larger declines. Under GPT-5.4, D1 decreases from $3.357$ to $3.270$, while D2, D3, and D4 decrease from $2.880$, $2.117$, and $1.743$ to $2.730$, $2.017$, and $1.543$, respectively. Under Opus, the corresponding changes are from $3.173$ to $3.020$ for D1 and from $2.467$, $2.220$, and $1.823$ to $2.273$, $2.053$, and $1.693$ for D2--D4. The largest change is in D4 under GPT-5.4, while Opus shows a broader reduction across the post-detection dimensions. D5 is comparatively stable in the aggregate, although its conflict-type trajectory differs because additional distractors provide traceable user details without improving diagnosis. Figure~\ref{fig:rq3_summary}(a,b) reports the aggregate trajectory and the dimension-level decreases on a positive-loss scale.

\textbf{The qualitative pattern is shared across conflict types, with different manifestations in the score profiles.} Across conflict types, the degradation is heterogeneous rather than uniform. In this paired cohort, CPC exhibits the clearest and most consistent decline in conflict-sensitive dimensions under both evaluations, particularly in diagnosis, calibration, and clarification; BOC shows a weaker D1--D4 decline with a partly separate faithfulness pattern, while SCC is comparatively stable. This variation is consistent with the structural demands of the axes: CPC requires the model to retain and use a context partition despite plausible same-domain alternatives. The matched-instance results support the aggregate interpretation: the evaluations are strongly associated overall (Pearson $r=0.853$), with lower agreement for BOC because its behavioral episodes require more interpretive judgments about whether a response sufficiently explains the underlying pattern. The complete type-level breakdown and case audit appear in Appendix~\ref{app:rq3_case_audit}.

\begin{figure*}[t]
    \centering
    \includegraphics[width=0.98\textwidth]{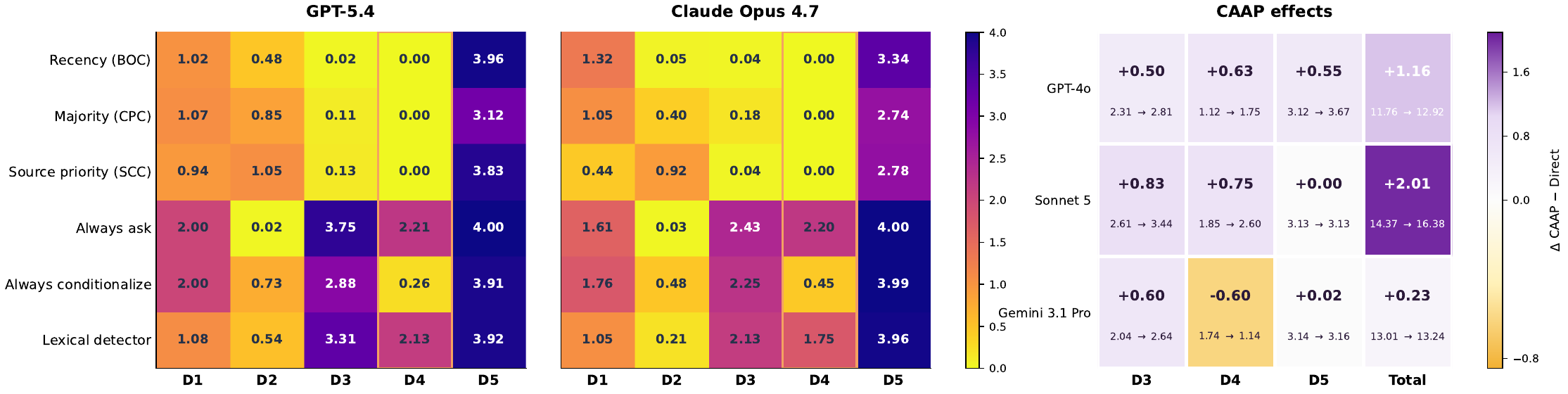}
    \caption{\textbf{Fixed policies fail structurally; CAAP improves behavior.} Left: fixed-policy profiles (0--4) under GPT-5.4 and Claude Opus 4.7. Right: CAAP-minus-direct-oracle changes under identical inputs, evaluated by GPT-5.4.}
    \label{fig:rq4_policy_tradeoff}
\end{figure*}
\subsection{Heuristic and Policy Baselines}
\label{sec:rq4}

Can simple heuristics resolve memory conflicts? We compare three axis-specific
value selectors---strict recency on 121 BOC instances, strict majority vote on
193 CPC instances, and fixed source priority on 227 SCC instances---with three
cross-axis conservative controls evaluated on all 541 instances: \emph{Ask-only},
which requests further information before recommending; \emph{Conditionalize-only},
which retains alternatives without selecting one; and a lexical conflict detector,
which uses surface opposition in the memory text to trigger a conservative
response.

The six policies introduced above serve two different diagnostic purposes.
The axis-specific selectors test whether an unresolved conflict is reduced
to a deterministic choice of behavioral phase, preference, or source, whereas
the cross-axis controls test simpler strategies that avoid commitment,
preserve alternatives, or react to surface-level opposition. The mechanism
statistics already show what is lost by these fixed rules: strict recency,
strict majority, and fixed source priority commit to one value on every
applicable instance, while majority voting is tied in 28.0\% of CPC
cases and has a margin of at most one in 71.5\% of cases. The lexical
detector likewise identifies only a small subset of conflict-positive banks
(recall 0.181), because surface opposition does not reliably expose temporal,
contextual, or source-dependent conflict. The selectors therefore serve as
structural ablations, and the controls as reference behaviors rather than
interchangeable policy competitors: the former discard conflict-bearing
relations, while the latter avoids or delays commitment without necessarily
explaining the underlying conflict. Figure~\ref{fig:rq4_policy_tradeoff}(a) summarizes these cross-judge failure profiles.
Because the rubric explicitly evaluates calibrated action, information
seeking, and memory faithfulness, the figures are descriptive rather than a
universal policy ranking.

We therefore design the \emph{Conflict-Aware Action Policy} (CAAP) to move
beyond fixed value selection and fixed response templates. Rather than always
selecting one value, always asking, or always conditionalizing, CAAP first decides
how to handle the conflict and then writes a response that follows that decision.
Given a visible memory bank $M$ and user query $q$, it uses two stages:

\[
(z,a) = \pi_{\theta}^{\mathrm{dec}}(M,q),
\qquad
r = \pi_{\theta}^{\mathrm{real}}(M,q,z,a).
\]
Here, $\pi^{\mathrm{dec}}$ simply denotes the first-stage decision module. It
produces a brief internal decision record $z$ (for example, what is uncertain
and what information is missing) and selects an action $a \in \mathcal{A}$.
$\pi^{\mathrm{real}}$ denotes the second-stage response-writing module: it uses
the visible memories, query, and selected action to produce the final response
$r$. Its action space is
\[
\mathcal{A} = \{\texttt{commit},\ \texttt{conditionalize},\ \texttt{clarify},\
\texttt{verify},\ \texttt{defer},\ \texttt{reversible\_trial}\}.
\]
Both stages use the same response-model backbone within each matched evaluation.
CAAP receives only the visible memory texts and the user query; it is not given
definition-level information such as the conflict type. The decision stage is instructed to choose the least cautious action that
is still justified by the visible evidence, while becoming more cautious when
an incorrect recommendation would have greater consequences. The realization
stage then turns that decision into a natural user-facing response.

We compare CAAP with direct oracle responses for Claude Sonnet~5,
Gemini~3.1~Pro, and GPT-4o while holding the response model, visible memory
bank, query, and Judge constant. Figure~\ref{fig:rq4_policy_tradeoff}(b) reports the matched CAAP comparison. CAAP is designed to select an appropriate conflict-handling action for each instance, so D3--D5 are the primary outcomes, while D1--D2 provide diagnostic context.

The three model evaluations show that CAAP's flexibility is useful but not
uniformly realized. Sonnet~5 improves in both action calibration and
decision-relevant coverage while preserving faithfulness, with the largest
gains on BOC and SCC. Gemini~3.1~Pro also improves action calibration, but
its coverage decreases, leaving little overall change. GPT-4o improves in
action calibration, clarification, and faithfulness under CAAP, although BOC
remains its weakest conflict type. Across the three models, CAAP supports
evidence-sensitive action selection, but the magnitude and composition of
its benefit depend on how reliably each response backbone realizes the
selected action and supplies the evidence required by the case.

CAAP action distributions provide a mechanism-level audit rather than the primary outcome: without explicit axis labels, it tends toward reversible trials for BOC, conditionalization or clarification for CPC, and verification or source-sensitive caution for SCC. The model-specific distributions and the full D1--D5 comparison are reported in Appendix~\ref{app:rq4_policy}. CAAP is therefore a positive action-policy baseline rather than an upper bound: it moves beyond a single fixed response template, while temporal-pattern reconstruction and detailed source reconciliation remain open challenges.

\section{Conclusion}
\label{sec:conclusion}

We introduced \textsc{TANGLE}, a benchmark for evaluating memory agents
under genuinely unresolvable conflict. Across 541 instances, five response
models, and four memory systems, we find a consistent recognition-to-action
gap: models detect and describe conflict more reliably than they calibrate
actions or seek clarification, while memory extraction and distractor noise
limit the conflict structure available for reasoning. Fixed policies further
compress or sidestep this structure, motivating CAAP as a flexible,
evidence-sensitive action policy. These findings suggest that reliable memory
agents needs to preserve conflicting evidence and its relations, represent what
remains unresolved, and choose actions that are calibrated to both the
available evidence and the consequences of error.

\bibliography{reference}
\bibliographystyle{dilab_ref}

\makeappendixtoc
\appendix
\clearpage
\onecolumn

\section{Persona Definitions}
\label{app:personas}

Starting from Persona Hub~\citep{personahub} seeds, we sample and curate 40 personas to cover a broad range of life circumstances. Persona Hub provides the initial profile material; it is not treated as a fixed set of benchmark instances. Because the source profiles do not have formal identifiers that we retain, the IDs in Table~\ref{tab:personas_full} are internal row identifiers assigned for this benchmark, not Persona Hub IDs. We normalize the sampled profiles into a common schema and lightly revise them where needed to make the life circumstances coherent and to support the benchmark's coverage objectives. The resulting personas span age (20--78), occupation (e.g., student, nurse, freelancer, warehouse associate, attorney, firefighter, retiree), income (low to high), family structure (single, roommates, partnered, single-parent, married-with-children, multi-generational, blended), health status (healthy, chronic conditions such as lupus, CKD, or diabetes, disabilities including wheelchair use and Deafness, and ADHD), and location (urban, suburban, rural, expatriate). Each persona is mapped to only 13--15 relevant attributes (of 46 distinct serialized names; 48 axis-specific labels; see \S\ref{app:attributes}), rather than the full persona\,$\times$\,attribute product; the complete persona--attribute mapping is maintained as part of the dataset metadata. Table~\ref{tab:personas_full} reports the normalized persona profiles and the life constraints used to ground their conflict instances, documenting the profile grounding and preserving plausible tensions rather than arbitrary contradictions.

{\small
\begin{xltabular}{\linewidth}{@{}l l Y Y Y@{}}
\caption{Persona roster and conflict-relevant grounding. IDs are internal benchmark identifiers (p01--p40), not source identifiers. We deliberately assign each persona only 13--15 life-relevant instances rather than the full persona$\times$attribute product. The final column records recurring pressures used to make conflicts ecologically plausible.}
\label{tab:personas_full}\\
\toprule
\textbf{ID} & \begin{tabular}[t]{@{}l@{}}\textbf{Name}\\textbf{Age}\end{tabular} & \textbf{Work / life stage} & \textbf{Household and setting} & \textbf{Conflict-relevant anchors} \\
\midrule
\endfirsthead
\multicolumn{5}{c}{\tablename~\thetable\ (continued)}\\
\toprule
\textbf{ID} & \begin{tabular}[t]{@{}l@{}}\textbf{Name}\\textbf{Age}\end{tabular} & \textbf{Work / life stage} & \textbf{Household and setting} & \textbf{Conflict-relevant anchors} \\
\midrule
\endhead
\midrule
\multicolumn{5}{r}{\footnotesize\emph{continued on next page}}\\
\endfoot
\bottomrule
\endlastfoot
\multicolumn{5}{@{}l}{\cellcolor[gray]{0.92}\textbf{Early career, education, and entry-level work}}\\
p01 & \begin{tabular}[t]{@{}l@{}}Maya\\(19--22)\end{tabular} & College student and barista & Three roommates in a city & Irregular work and class schedule; tight budget; shared-space coordination \\
p02 & \begin{tabular}[t]{@{}l@{}}Jordan\\(24--29)\end{tabular} & Junior software engineer, hybrid & Lives alone in an expensive city & Cross-time-zone collaboration; sustained cognitive load; ``always-on'' work expectations \\
p03 & \begin{tabular}[t]{@{}l@{}}Elena\\(31--38)\end{tabular} & Registered nurse on rotating shifts & Partner and young child & Night and weekend shifts; sleep disruption; childcare handoffs \\
p04 & \begin{tabular}[t]{@{}l@{}}DeShawn\\(34--42)\end{tabular} & Warehouse associate on shifts & Multi-generational household & Physically demanding work; little schedule control; financial support for family \\
p05 & \begin{tabular}[t]{@{}l@{}}Priya\\(36--44)\end{tabular} & Middle-school teacher & Married with two children & Evening spillover from work; emotional labor; rigid school-day schedule \\
p06 & \begin{tabular}[t]{@{}l@{}}Luis\\(28--37)\end{tabular} & Freelance graphic designer & Works from home with partner & Volatile income; client deadlines; self-managed work boundaries \\
p07 & \begin{tabular}[t]{@{}l@{}}Amina\\(40--50)\end{tabular} & Small grocery-store owner & Spouse and extended family & Long operating hours; staffing shortfalls; intertwined household and business finances \\
p08 & \begin{tabular}[t]{@{}l@{}}Noah\\(45--55)\end{tabular} & Mid-level corporate manager & Spouse and eldercare responsibilities & Competing work and caregiving demands; medical coordination; decision fatigue \\
p09 & \begin{tabular}[t]{@{}l@{}}Grace\\(52--60)\end{tabular} & Public-transit bus driver & Lives alone & Fixed routes and breaks; minimal flexibility; strict punctuality requirements \\
p10 & \begin{tabular}[t]{@{}l@{}}Ethan\\(61--70)\end{tabular} & Recently retired accountant & Spouse in a suburb & Identity transition after retirement; fixed income; rising medical needs \\
\multicolumn{5}{@{}l}{\cellcolor[gray]{0.92}\textbf{Precarious, care-intensive, and high-demand work}}\\
p11 & \begin{tabular}[t]{@{}l@{}}Sofia\\(27--35)\end{tabular} & Hospitality worker and immigrant single parent & Relatives and one child & Language friction; variable shifts; limited childcare options \\
p12 & \begin{tabular}[t]{@{}l@{}}Caleb\\(30--40)\end{tabular} & Gig driver and online learner & Lives with a roommate & Algorithm-driven income; no paid leave; self-funded reskilling \\
p13 & \begin{tabular}[t]{@{}l@{}}Riley\\(18--21)\end{tabular} & Retail cashier and community-college student & Single parent with siblings in a small town & Unreliable transit; contributes to bills; limited study space \\
p14 & \begin{tabular}[t]{@{}l@{}}Owen\\(22--27)\end{tabular} & Line cook & Shares an apartment with coworkers & Late-night shifts; volatile tips; sleep and noise disruption \\
p15 & \begin{tabular}[t]{@{}l@{}}Zoe\\(23--30)\end{tabular} & Social-media coordinator at a startup & Partner in an expensive city & Always-on culture; high rent; engagement-metric pressure \\
p16 & \begin{tabular}[t]{@{}l@{}}Malik\\(26--33)\end{tabular} & Paramedic / first responder & Fianc\'e, infant, and outer-suburban home & Trauma exposure; unpredictable overtime; childcare gaps \\
p17 & \begin{tabular}[t]{@{}l@{}}Hannah\\(29--36)\end{tabular} & Attorney at a large firm & Married, no children, downtown & Billable-hour pressure; frequent travel; financial and practical support for a parent \\
p18 & \begin{tabular}[t]{@{}l@{}}Victor\\(32--40)\end{tabular} & Union electrician & Co-parents one child across two households & Weather-sensitive jobs; custody coordination; injury risk \\
p19 & \begin{tabular}[t]{@{}l@{}}Leah\\(35--43)\end{tabular} & HR director & Divorced, two children, suburban & Dual-household logistics; workplace conflict; little personal time \\
p20 & \begin{tabular}[t]{@{}l@{}}Samir\\(38--46)\end{tabular} & Long-haul truck driver & Spouse in a rural area & Long absences; chronic back pain; limited healthcare access on the road \\
p21 & \begin{tabular}[t]{@{}l@{}}Nadia\\(41--49)\end{tabular} & City police sergeant & Single; caregiver for an elderly aunt & Rotating shifts; public scrutiny; caregiver burnout \\
p22 & \begin{tabular}[t]{@{}l@{}}Trevor\\(44--52)\end{tabular} & Regional sales executive & Remarried, blended family & Heavy travel; stepfamily tensions; quota-linked compensation \\
p23 & \begin{tabular}[t]{@{}l@{}}Mei\\(47--55)\end{tabular} & Home-health aide & Rents a room in a shared house & Low wages; emotional labor; remittance obligations \\
p24 & \begin{tabular}[t]{@{}l@{}}Andre\\(50--58)\end{tabular} & Manufacturing-plant supervisor & Married; adult son has moved home & Sandwich-generation demands; production deadlines; hypertension management \\
\multicolumn{5}{@{}l}{\cellcolor[gray]{0.92}\textbf{Later career, retirement transition, and health-related constraints}}\\
p25 & \begin{tabular}[t]{@{}l@{}}Kendra\\(53--61)\end{tabular} & State corrections officer & Lives alone in an exurban area & High-stress safety work; long commute; limited local support \\
p26 & \begin{tabular}[t]{@{}l@{}}Farah\\(55--63)\end{tabular} & Pharmacist & Married; helps with family business & Medication liability; long periods standing; competing household and business finances \\
p27 & \begin{tabular}[t]{@{}l@{}}Ben\\(58--66)\end{tabular} & Farmer & Spouse in a rural, multi-generational household & Weather volatility; equipment debt; limited local services \\
p28 & \begin{tabular}[t]{@{}l@{}}Carla\\(60--68)\end{tabular} & Recently widowed hotel housekeeper & Senior apartment & Fixed income; grief and isolation; physically demanding work \\
p29 & \begin{tabular}[t]{@{}l@{}}Dmitri\\(62--70)\end{tabular} & Post-retirement rideshare driver and immigrant & Spouse & Language barriers; variable earnings; diabetes management \\
p30 & \begin{tabular}[t]{@{}l@{}}Janice\\(64--72)\end{tabular} & Part-time librarian & Caregiver for spouse with dementia & Caregiver fatigue; medical-system navigation; reduced social contact \\
p31 & \begin{tabular}[t]{@{}l@{}}Haruto\\(67--75)\end{tabular} & Retired engineer and expatriate & Splits time across two countries & Visa and tax complexity; fragmented health records; cross-border coordination \\
p32 & \begin{tabular}[t]{@{}l@{}}Lila\\(70--78)\end{tabular} & Retired teacher and wheelchair user & Alone in an older building & Accessibility barriers; fixed income; dependence on paratransit \\
\multicolumn{5}{@{}l}{\cellcolor[gray]{0.92}\textbf{Disability, chronic illness, mobility, and nonstandard family arrangements}}\\
p33 & \begin{tabular}[t]{@{}l@{}}Miguel\\(24--32)\end{tabular} & Deaf customer-support specialist, remote & Lives with partner & Inaccessible meetings; captioning failures; career bias \\
p34 & \begin{tabular}[t]{@{}l@{}}Avery\\(27--34)\end{tabular} & UX researcher with ADHD & Co-lives with two friends & Executive-function demands; time blindness; meeting overload \\
p35 & \begin{tabular}[t]{@{}l@{}}Rosa\\(33--41)\end{tabular} & Municipal clerk with lupus & Single parent of one & Unpredictable flare-ups; strict attendance rules; fragile childcare arrangements \\
p36 & \begin{tabular}[t]{@{}l@{}}Imani\\(30--38)\end{tabular} & Army reservist and logistics coordinator & Married with a toddler & Deployment interruptions; spouse's career sacrifices; childcare continuity \\
p37 & \begin{tabular}[t]{@{}l@{}}Chen\\(39--47)\end{tabular} & International-school teacher and expatriate & Abroad with spouse and teenager & Visa insecurity; cultural friction; distant eldercare \\
p38 & \begin{tabular}[t]{@{}l@{}}Paula\\(42--50)\end{tabular} & Restaurant-franchise owner & Married; supports a college-age child & Debt and cash-flow pressure; staff turnover; weekend and holiday work \\
p39 & \begin{tabular}[t]{@{}l@{}}Greg\\(46--54)\end{tabular} & Seasonal wildfire firefighter & Rural rental with partner & Dangerous seasonal surges; off-season instability; cumulative health effects \\
p40 & \begin{tabular}[t]{@{}l@{}}Selene\\(57--65)\end{tabular} & Clinical-trial participant with CKD; part-time bookkeeper & Lives with sister & Frequent treatment; fatigue; insurance uncertainty \\
\end{xltabular}
}

\section{Attribute Schema}
\label{app:attributes}

Table~\ref{tab:attributes_full} lists the attributes used in the benchmark, organized by domain and conflict axis.

{\footnotesize
\begin{xltabular}{\linewidth}{@{}Y c Y@{}}
\caption{All 46 attributes grouped by life domain. Axis indicates the conflict type the attribute is designed to elicit (BOC = Behavior-Oscillation, CPC = Context-Partitioned, SCC = Source-Contradiction).}
\label{tab:attributes_full}\\
\toprule
\textbf{Attribute} & \textbf{Axis} & \textbf{Conflict elicited} \\
\midrule
\endfirsthead
\multicolumn{3}{c}{\tablename~\thetable\ (continued)}\\
\toprule
\textbf{Attribute} & \textbf{Axis} & \textbf{Conflict elicited} \\
\midrule
\endhead
\midrule
\multicolumn{3}{r}{\footnotesize\emph{continued on next page}}\\
\endfoot
\bottomrule
\endlastfoot
\multicolumn{3}{@{}l}{\cellcolor[gray]{0.92}\textbf{Health}}\\
\texttt{exercise\_pattern} & BOC & Workout type/frequency/timing oscillates \\
\texttt{sleep\_schedule} & BOC & Sleep/wake patterns shift with life changes \\
\texttt{meal\_prep} & BOC & Approach to cooking/planning meals oscillates \\
\texttt{dietary\_trigger} & CPC & Food choices differ by context (body state, rules, audience) \\
\texttt{dietary\_choice} & CPC & Eating philosophy shifts by goal phase or context \\
\texttt{substance\_moderation} & BOC, CPC & Alcohol/substance behavior oscillates or rules differ by social visibility or risk \\
\multicolumn{3}{@{}l}{\cellcolor[gray]{0.92}\textbf{Finance}}\\
\texttt{budget\_tracking} & BOC & Tracking method/rigor oscillates \\
\texttt{spending\_discipline} & BOC & Spending control oscillates \\
\texttt{financial\_record} & SCC & Sources disagree on account balances/transactions \\
\texttt{coverage\_eligibility} & SCC & Sources disagree on insurance/benefit eligibility \\
\texttt{insurance\_claim\_evidence} & SCC & Sources disagree on claim documentation \\
\multicolumn{3}{@{}l}{\cellcolor[gray]{0.92}\textbf{Work}}\\
\texttt{career\_commitment} & BOC & Commitment to current career path oscillates \\
\texttt{career\_commitment\_cpc} & CPC & Career priorities differ by relational/resource context \\
\texttt{email\_triage} & BOC & Email management approach oscillates \\
\texttt{scheduling\_policy} & CPC & Scheduling rules differ by physiological/risk/resource context \\
\texttt{communication\_style} & CPC & Tone/detail level differs by audience/commitment \\
\texttt{work\_communication\_mode} & CPC & Channel choice (Slack/email/call) differs by risk/audience \\
\texttt{boundary\_enforcement} & CPC & Boundary strictness differs by audience/risk/rule context \\
\texttt{information\_consumption} & CPC & Reading depth/format differs by social visibility/goal phase \\
\multicolumn{3}{@{}l}{\cellcolor[gray]{0.92}\textbf{Routine}}\\
\texttt{training\_structure} & BOC & Workout structure/planning oscillates \\
\texttt{knowledge\_capture} & BOC & Note-taking/learning system oscillates \\
\texttt{skill\_learning} & BOC & Approach to learning new skills oscillates \\
\texttt{household\_commitment\_drift} & BOC & Home/care responsibility commitment oscillates \\
\multicolumn{3}{@{}l}{\cellcolor[gray]{0.92}\textbf{Social}}\\
\texttt{friendship\_maintenance} & BOC & Social connection effort oscillates \\
\texttt{social\_engagement} & CPC & Social participation differs by resource/energy context \\
\texttt{venue\_preference} & CPC & Venue choice differs by commitment level \\
\multicolumn{3}{@{}l}{\cellcolor[gray]{0.92}\textbf{Mobility}}\\
\texttt{transport\_preference} & CPC & Transport mode differs by location/risk/physiological state \\
\texttt{travel\_standard} & CPC & Travel comfort/budget tradeoff differs by context \\
\texttt{evacuation\_trigger} & CPC & Evacuation decision threshold differs by competence/rule context \\
\multicolumn{3}{@{}l}{\cellcolor[gray]{0.92}\textbf{Consumption}}\\
\texttt{purchase\_criteria} & CPC & Buying priorities differ by beneficiary/risk/social visibility \\
\texttt{household\_procurement} & CPC & Shopping approach differs by audience/risk context \\
\texttt{home\_access\_credential} & CPC & Access-sharing rules differ by audience \\
\multicolumn{3}{@{}l}{\cellcolor[gray]{0.92}\textbf{Crisis}}\\
\texttt{crisis\_resource\_allocation} & BOC & Resource allocation under pressure oscillates \\
\multicolumn{3}{@{}l}{\cellcolor[gray]{0.92}\textbf{Family}}\\
\texttt{care\_decision\_authority} & SCC & Sources disagree on who has medical/care decision authority \\
\texttt{co\_parent\_custody} & SCC & Sources disagree on custody arrangements \\
\texttt{dependent\_benefit} & SCC & Sources disagree on benefit eligibility for dependents \\
\texttt{will\_beneficiary} & SCC & Sources disagree on estate/beneficiary designations \\
\multicolumn{3}{@{}l}{\cellcolor[gray]{0.92}\textbf{Admin}}\\
\texttt{document\_deadline} & SCC & Sources disagree on document due dates \\
\texttt{account\_recovery} & SCC & Sources disagree on account ownership/verification \\
\texttt{authorization\_status} & SCC & Sources disagree on authorization/permission status \\
\texttt{service\_status} & SCC & Sources disagree on service activation/cancellation \\
\texttt{delivery\_status} & SCC & Sources disagree on delivery/shipment status \\
\texttt{event\_scheduling} & SCC & Sources disagree on event time/location/details \\
\texttt{landlord\_tenant} & SCC & Sources disagree on lease terms/responsibilities \\
\texttt{medical\_record} & SCC & Sources disagree on medical history/diagnoses \\
\texttt{medical\_instruction} & SCC & Sources disagree on treatment instructions \\
\end{xltabular}
}

\section{Conflict-Structure Schema}
\label{app:conflict_structure_schema}

Whereas Table~\ref{tab:attributes_full} specifies the life aspect associated with an instance, the following tables specify the structural variable that makes its memories incompatible. Counts are computed from the 541 benchmark instances.

{\footnotesize
\begin{xltabular}{\linewidth}{@{}l r Y Y@{}}
\caption{CPC context-partition schema. The partition type identifies the contextual variable that determines which preference or behavior applies.}
\label{tab:cpc_partition_schema}\\
\toprule
\textbf{Partition type} & \textbf{Count} & \textbf{Variable to determine} & \textbf{Illustrative distinction} \\
\midrule
\endfirsthead
\multicolumn{4}{c}{\tablename~\thetable\ (continued)}\\
\toprule
\textbf{Partition type} & \textbf{Count} & \textbf{Variable to determine} & \textbf{Illustrative distinction} \\
\midrule
\endhead
\midrule
\multicolumn{4}{r}{\footnotesize\emph{continued on next page}}\\
\endfoot
\bottomrule
\endlastfoot
\texttt{audience} & 25 & Who the user is communicating with or acting for & formal with a supervisor vs. casual with peers \\
\texttt{risk} & 24 & How consequential the situation is & cautious for medical decisions vs. relaxed for routine choices \\
\texttt{goal\_phase} & 21 & Whether the user is exploring or executing & broad research vs. focused implementation \\
\texttt{resource} & 17 & Whether resources are abundant or scarce & splurge when flush vs. conserve when constrained \\
\texttt{rule\_governance} & 16 & Whether external rules constrain the action & strict under audit vs. flexible otherwise \\
\texttt{social\_visibility} & 13 & Whether the action is public or private & polished in public vs. informal in private \\
\texttt{relational} & 13 & Whether the action is for self or others & frugal for self vs. generous for children \\
\texttt{commitment\_level} & 13 & How deeply invested the user is & thorough for passion projects vs. minimal for routine tasks \\
\texttt{physiological} & 12 & The current physical state & avoids meetings during migraines \\
\texttt{competence} & 12 & Whether the domain is familiar & confident as an expert vs. cautious as a novice \\
\texttt{beneficiary} & 8 & Who benefits from the action & quality for work equipment vs. economy for household purchases \\
\texttt{location} & 7 & Where the user is acting & bikes nearby vs. takes transit to unfamiliar areas \\
\texttt{emotional\_state} & 3 & The current emotional state & commits during momentum vs. explores during doubt \\
\texttt{temporal} & 3 & The time-related setting & deep reading in the morning vs. headlines in the evening \\
\texttt{formality\_context} & 2 & Whether the setting is formal & structured in meetings vs. loose in chat \\
\texttt{power\_dynamics} & 2 & The power relationship between parties & formal with superiors vs. casual with peers \\
\texttt{energy\_state} & 2 & The current energy or depletion level & hosts when rested vs. withdraws when drained \\
\end{xltabular}
}

{\footnotesize
\begin{xltabular}{\linewidth}{@{}l r Y@{}}
\caption{BOC oscillation-driver schema. Drivers characterize the mechanism that sustains or triggers behavioral reversals.}
\label{tab:boc_driver_schema}\\
\toprule
\textbf{Driver} & \textbf{Count} & \textbf{Mechanism} \\
\midrule
\endfirsthead
\multicolumn{3}{c}{\tablename~\thetable\ (continued)}\\
\toprule
\textbf{Driver} & \textbf{Count} & \textbf{Mechanism} \\
\midrule
\endhead
\midrule
\multicolumn{3}{r}{\footnotesize\emph{continued on next page}}\\
\endfoot
\bottomrule
\endlastfoot
\texttt{identity\_conflict} & 16 & Competing self-concepts make neither behavioral pole simply erroneous. \\
\texttt{normative\_friction} & 16 & An effective method conflicts with the user's values or preferred self-image. \\
\texttt{environmental\_dependency} & 15 & A method requires conditions or infrastructure that do not persist. \\
\texttt{novelty\_stability} & 11 & Initial enthusiasm for a new tool or routine fades over time. \\
\texttt{reward\_horizon} & 11 & Short-term and delayed rewards favor different behavioral choices. \\
\texttt{social\_exposure} & 9 & Behavior differs when it is observed by others. \\
\texttt{threshold\_triggered} & 9 & Action changes only when a metric crosses a salient threshold. \\
\texttt{trust\_calibration} & 7 & Trust in a method fluctuates after successes or failures. \\
\texttt{relationship\_dynamics} & 7 & Family or partner involvement alternates between delegation and control. \\
\texttt{metric\_reactivity} & 7 & Tracking initially helps but later creates pressure or obsession. \\
\texttt{control\_delegation} & 6 & The user alternates between autonomy and cognitive offloading. \\
\texttt{outcome\_attribution} & 6 & Coincidental outcomes are credited to or blamed on the method. \\
\texttt{social\_pressure\_and\_exception\_rationalization} & 1 & Social pressure and exceptions destabilize an existing commitment. \\
\end{xltabular}
}

{\footnotesize
\begin{xltabular}{\linewidth}{@{}l r Y@{}}
\caption{BOC oscillation-shape schema. Shapes characterize the temporal form of the observed reversals.}
\label{tab:boc_shape_schema}\\
\toprule
\textbf{Shape} & \textbf{Count} & \textbf{Description} \\
\midrule
\endfirsthead
\multicolumn{3}{c}{\tablename~\thetable\ (continued)}\\
\toprule
\textbf{Shape} & \textbf{Count} & \textbf{Description} \\
\midrule
\endhead
\midrule
\multicolumn{3}{r}{\footnotesize\emph{continued on next page}}\\
\endfoot
\bottomrule
\endlastfoot
\texttt{method\_alternation} & 30 & Repeated switching between two methods. \\
\texttt{partial\_retention} & 22 & A core habit remains while peripheral practices oscillate. \\
\texttt{threshold\_toggling} & 20 & The approach changes when a metric crosses a threshold. \\
\texttt{burst\_decay} & 16 & An intense start is followed by gradual fading. \\
\texttt{amplitude\_drift} & 13 & The magnitude of the oscillation changes over time. \\
\texttt{escalation\_ladder} & 12 & Mild behavior escalates, overcorrects, and then collapses. \\
\texttt{channel\_substitution} & 7 & The goal remains stable while the method or channel rotates. \\
\texttt{commit\_succeed\_trigger\_relapse\_recommit} & 1 & Commitment succeeds, a trigger causes relapse, and commitment then restarts. \\
\end{xltabular}
}

{\footnotesize
\begin{xltabular}{\linewidth}{@{}l r Y@{}}
\caption{SCC source-conflict schema. Types characterize the form of disagreement among sources.}
\label{tab:scc_type_schema}\\
\toprule
\textbf{Source-conflict type} & \textbf{Count} & \textbf{Description} \\
\midrule
\endfirsthead
\multicolumn{3}{c}{\tablename~\thetable\ (continued)}\\
\toprule
\textbf{Source-conflict type} & \textbf{Count} & \textbf{Description} \\
\midrule
\endhead
\midrule
\multicolumn{3}{r}{\footnotesize\emph{continued on next page}}\\
\endfoot
\bottomrule
\endlastfoot
\texttt{interpretation\_conflict} & 49 & Sources interpret the same evidence or rule differently. \\
\texttt{record\_vs\_reality} & 33 & An official record conflicts with the user's actual situation. \\
\texttt{policy\_vs\_exception} & 30 & A general policy conflicts with an exception or waiver. \\
\texttt{multi\_source} & 30 & Three or more sources disagree without a single clear resolution path. \\
\texttt{evidence\_quality} & 24 & Sources differ in the quality or medium of their evidence. \\
\texttt{jurisdiction\_collision} & 20 & Overlapping authorities issue incompatible instructions. \\
\texttt{identity\_linkage\_error} & 15 & Information is attributed to the wrong person or entity. \\
\texttt{temporal\_inconsistency} & 13 & The same source gives incompatible statements at different times. \\
\texttt{incentive\_biased} & 8 & A source's incentives may bias its claim. \\
\texttt{legal\_hierarchy} & 5 & Different levels of rules or regulations conflict. \\
\end{xltabular}
}

SCC instances contain two, three, or four conflicting sources (122, 83, and 22 instances, respectively).

\section{Persona--Attribute Coverage}
\label{app:persona_attributes}

The following tables detail the sparse persona--attribute coverage.

{\footnotesize
\begin{xltabular}{\linewidth}{@{}l l r r r r @ {\hspace{\tabcolsep}\vrule width 0.4pt\hspace{\tabcolsep}} l l r r r r@{}}
\caption{Per-persona counts in the sparse persona--attribute assignment. IDs are internal benchmark identifiers.}
\label{tab:persona_attribute_counts}\\
\toprule
\textbf{ID} & \textbf{Name} & \textbf{\#} & \textbf{BOC} & \textbf{CPC} & \textbf{SCC} & \textbf{ID} & \textbf{Name} & \textbf{\#} & \textbf{BOC} & \textbf{CPC} & \textbf{SCC} \\
\midrule
\endfirsthead
\multicolumn{12}{c}{\tablename~\thetable\ (continued)}\\
\toprule
\textbf{ID} & \textbf{Name} & \textbf{\#} & \textbf{BOC} & \textbf{CPC} & \textbf{SCC} & \textbf{ID} & \textbf{Name} & \textbf{\#} & \textbf{BOC} & \textbf{CPC} & \textbf{SCC} \\
\midrule
\endhead
\midrule
\multicolumn{12}{r}{\footnotesize\emph{continued on next page}}\\
\endfoot
\bottomrule
\endlastfoot
p01 & Maya & 14 & 8 & 2 & 4 & p21 & Nadia & 13 & 1 & 4 & 8 \\
p02 & Jordan & 14 & 7 & 4 & 3 & p22 & Trevor & 13 & 2 & 6 & 5 \\
p03 & Elena & 14 & 3 & 5 & 6 & p23 & Mei & 13 & 1 & 4 & 8 \\
p04 & DeShawn & 14 & 4 & 6 & 4 & p24 & Andre & 13 & 4 & 3 & 6 \\
p05 & Priya & 14 & 3 & 6 & 5 & p25 & Kendra & 13 & 2 & 5 & 6 \\
p06 & Luis & 13 & 4 & 6 & 3 & p26 & Farah & 15 & 1 & 5 & 9 \\
p07 & Amina & 14 & 4 & 6 & 4 & p27 & Ben & 13 & 4 & 4 & 5 \\
p08 & Noah & 14 & 4 & 5 & 5 & p28 & Carla & 13 & 3 & 4 & 6 \\
p09 & Grace & 14 & 2 & 7 & 5 & p29 & Dmitri & 13 & 1 & 6 & 6 \\
p10 & Ethan & 14 & 4 & 4 & 6 & p30 & Janice & 13 & 3 & 1 & 9 \\
p11 & Sofia & 14 & 3 & 4 & 7 & p31 & Haruto & 13 & 2 & 6 & 5 \\
p12 & Caleb & 14 & 4 & 5 & 5 & p32 & Lila & 13 & 1 & 3 & 9 \\
p13 & Riley & 13 & 6 & 4 & 3 & p33 & Miguel & 14 & 2 & 8 & 4 \\
p14 & Owen & 14 & 4 & 6 & 4 & p34 & Avery & 13 & 5 & 6 & 2 \\
p15 & Zoe & 14 & 5 & 6 & 3 & p35 & Rosa & 13 & 1 & 3 & 9 \\
p16 & Malik & 14 & 3 & 4 & 7 & p36 & Imani & 13 & 5 & 4 & 4 \\
p17 & Hannah & 14 & 2 & 7 & 5 & p37 & Chen & 13 & 2 & 5 & 6 \\
p18 & Victor & 14 & 1 & 6 & 7 & p38 & Paula & 13 & 4 & 4 & 5 \\
p19 & Leah & 14 & 0 & 5 & 9 & p39 & Greg & 13 & 2 & 6 & 5 \\
p20 & Samir & 14 & 3 & 6 & 5 & p40 & Selene & 13 & 1 & 2 & 10 \\
\end{xltabular}
}

{\scriptsize
\begin{xltabular}{\linewidth}{@{}l l Y Y Y@{}}
\caption{\footnotesize Detailed persona--attribute assignment by conflict axis. Attributes are shown using the names defined in Table~\ref{tab:attributes_full}.}
\label{tab:persona_attribute_map}\\
\toprule
{\footnotesize\textbf{ID}} & {\footnotesize\textbf{Name}} & {\footnotesize\textbf{BOC attributes}} & {\footnotesize\textbf{CPC attributes}} & {\footnotesize\textbf{SCC attributes}} \\
\midrule
\endfirsthead
\multicolumn{5}{c}{\tablename~\thetable\ (continued)}\\
\toprule
{\footnotesize\textbf{ID}} & {\footnotesize\textbf{Name}} & {\footnotesize\textbf{BOC attributes}} & {\footnotesize\textbf{CPC attributes}} & {\footnotesize\textbf{SCC attributes}} \\
\midrule
\endhead
\midrule
\multicolumn{5}{r}{\footnotesize\emph{continued on next page}}\\
\endfoot
\bottomrule
\endlastfoot
p01 & Maya & \texttt{budget\_tracking}, \texttt{friendship\_maintenance}, \texttt{knowledge\_capture}, \texttt{meal\_prep}, \texttt{skill\_learning}, \texttt{sleep\_schedule}, \texttt{spending\_discipline}, \texttt{training\_structure} & \texttt{transport\_preference}, \texttt{work\_communication\_mode} & \texttt{account\_recovery}, \texttt{document\_deadline}, \texttt{event\_scheduling}, \texttt{financial\_record} \\
p02 & Jordan & \texttt{career\_commitment}, \texttt{email\_triage}, \texttt{exercise\_pattern}, \texttt{knowledge\_capture}, \texttt{skill\_learning}, \texttt{sleep\_schedule}, \texttt{training\_structure} & \texttt{boundary\_enforcement}, \texttt{information\_consumption}, \texttt{scheduling\_policy}, \texttt{work\_communication\_mode} & \texttt{account\_recovery}, \texttt{authorization\_status}, \texttt{insurance\_claim\_evidence} \\
p03 & Elena & \texttt{crisis\_resource\_allocation}, \texttt{substance\_moderation}, \texttt{training\_structure} & \texttt{boundary\_enforcement}, \texttt{career\_commitment\_cpc}, \texttt{communication\_style}, \texttt{evacuation\_trigger}, \texttt{work\_communication\_mode} & \texttt{authorization\_status}, \texttt{care\_decision\_authority}, \texttt{document\_deadline}, \texttt{medical\_instruction}, \texttt{medical\_record}, \texttt{service\_status} \\
p04 & DeShawn & \texttt{budget\_tracking}, \texttt{exercise\_pattern}, \texttt{household\_commitment\_drift}, \texttt{meal\_prep} & \texttt{dietary\_choice}, \texttt{household\_procurement}, \texttt{purchase\_criteria}, \texttt{scheduling\_policy}, \texttt{substance\_moderation}, \texttt{transport\_preference} & \texttt{delivery\_status}, \texttt{financial\_record}, \texttt{landlord\_tenant}, \texttt{service\_status} \\
p05 & Priya & \texttt{friendship\_maintenance}, \texttt{skill\_learning}, \texttt{training\_structure} & \texttt{communication\_style}, \texttt{dietary\_trigger}, \texttt{household\_procurement}, \texttt{information\_consumption}, \texttt{social\_engagement}, \texttt{venue\_preference} & \texttt{authorization\_status}, \texttt{care\_decision\_authority}, \texttt{dependent\_benefit}, \texttt{document\_deadline}, \texttt{event\_scheduling} \\
p06 & Luis & \texttt{budget\_tracking}, \texttt{email\_triage}, \texttt{knowledge\_capture}, \texttt{spending\_discipline} & \texttt{career\_commitment\_cpc}, \texttt{information\_consumption}, \texttt{purchase\_criteria}, \texttt{scheduling\_policy}, \texttt{substance\_moderation}, \texttt{travel\_standard} & \texttt{account\_recovery}, \texttt{financial\_record}, \texttt{insurance\_claim\_evidence} \\
p07 & Amina & \texttt{budget\_tracking}, \texttt{career\_commitment}, \texttt{crisis\_resource\_allocation}, \texttt{meal\_prep} & \texttt{boundary\_enforcement}, \texttt{career\_commitment}, \texttt{home\_access\_credential}, \texttt{household\_procurement}, \texttt{purchase\_criteria}, \texttt{work\_communication\_mode} & \texttt{account\_recovery}, \texttt{delivery\_status}, \texttt{financial\_record}, \texttt{service\_status} \\
p08 & Noah & \texttt{crisis\_resource\_allocation}, \texttt{friendship\_maintenance}, \texttt{household\_commitment\_drift}, \texttt{meal\_prep} & \texttt{boundary\_enforcement}, \texttt{communication\_style}, \texttt{scheduling\_policy}, \texttt{social\_engagement}, \texttt{transport\_preference} & \texttt{authorization\_status}, \texttt{care\_decision\_authority}, \texttt{dependent\_benefit}, \texttt{event\_scheduling}, \texttt{medical\_instruction} \\
p09 & Grace & \texttt{meal\_prep}, \texttt{sleep\_schedule} & \texttt{boundary\_enforcement}, \texttt{career\_commitment\_cpc}, \texttt{communication\_style}, \texttt{dietary\_trigger}, \texttt{evacuation\_trigger}, \texttt{substance\_moderation}, \texttt{transport\_preference} & \texttt{financial\_record}, \texttt{insurance\_claim\_evidence}, \texttt{landlord\_tenant}, \texttt{medical\_record}, \texttt{service\_status} \\
p10 & Ethan & \texttt{budget\_tracking}, \texttt{friendship\_maintenance}, \texttt{sleep\_schedule}, \texttt{spending\_discipline} & \texttt{information\_consumption}, \texttt{social\_engagement}, \texttt{travel\_standard}, \texttt{venue\_preference} & \texttt{account\_recovery}, \texttt{coverage\_eligibility}, \texttt{dependent\_benefit}, \texttt{event\_scheduling}, \texttt{medical\_record}, \texttt{will\_beneficiary} \\
p11 & Sofia & \texttt{budget\_tracking}, \texttt{meal\_prep}, \texttt{spending\_discipline} & \texttt{dietary\_choice}, \texttt{dietary\_trigger}, \texttt{home\_access\_credential}, \texttt{household\_procurement} & \texttt{authorization\_status}, \texttt{care\_decision\_authority}, \texttt{coverage\_eligibility}, \texttt{delivery\_status}, \texttt{dependent\_benefit}, \texttt{document\_deadline}, \texttt{landlord\_tenant} \\
p12 & Caleb & \texttt{career\_commitment}, \texttt{skill\_learning}, \texttt{spending\_discipline}, \texttt{training\_structure} & \texttt{career\_commitment}, \texttt{information\_consumption}, \texttt{purchase\_criteria}, \texttt{scheduling\_policy}, \texttt{transport\_preference} & \texttt{account\_recovery}, \texttt{co\_parent\_custody}, \texttt{document\_deadline}, \texttt{financial\_record}, \texttt{service\_status} \\
p13 & Riley & \texttt{budget\_tracking}, \texttt{friendship\_maintenance}, \texttt{meal\_prep}, \texttt{skill\_learning}, \texttt{spending\_discipline}, \texttt{training\_structure} & \texttt{dietary\_choice}, \texttt{household\_procurement}, \texttt{scheduling\_policy}, \texttt{social\_engagement} & \texttt{account\_recovery}, \texttt{delivery\_status}, \texttt{event\_scheduling} \\
p14 & Owen & \texttt{exercise\_pattern}, \texttt{meal\_prep}, \texttt{sleep\_schedule}, \texttt{spending\_discipline} & \texttt{dietary\_choice}, \texttt{dietary\_trigger}, \texttt{scheduling\_policy}, \texttt{substance\_moderation}, \texttt{transport\_preference}, \texttt{work\_communication\_mode} & \texttt{financial\_record}, \texttt{insurance\_claim\_evidence}, \texttt{medical\_instruction}, \texttt{service\_status} \\
p15 & Zoe & \texttt{career\_commitment}, \texttt{crisis\_resource\_allocation}, \texttt{email\_triage}, \texttt{knowledge\_capture}, \texttt{skill\_learning} & \texttt{boundary\_enforcement}, \texttt{career\_commitment}, \texttt{dietary\_trigger}, \texttt{purchase\_criteria}, \texttt{travel\_standard}, \texttt{work\_communication\_mode} & \texttt{account\_recovery}, \texttt{co\_parent\_custody}, \texttt{financial\_record} \\
p16 & Malik & \texttt{crisis\_resource\_allocation}, \texttt{friendship\_maintenance}, \texttt{sleep\_schedule} & \texttt{communication\_style}, \texttt{evacuation\_trigger}, \texttt{substance\_moderation}, \texttt{work\_communication\_mode} & \texttt{authorization\_status}, \texttt{care\_decision\_authority}, \texttt{document\_deadline}, \texttt{insurance\_claim\_evidence}, \texttt{medical\_instruction}, \texttt{medical\_record}, \texttt{service\_status} \\
p17 & Hannah & \texttt{email\_triage}, \texttt{knowledge\_capture} & \texttt{boundary\_enforcement}, \texttt{communication\_style}, \texttt{dietary\_choice}, \texttt{information\_consumption}, \texttt{scheduling\_policy}, \texttt{venue\_preference}, \texttt{work\_communication\_mode} & \texttt{account\_recovery}, \texttt{authorization\_status}, \texttt{care\_decision\_authority}, \texttt{document\_deadline}, \texttt{will\_beneficiary} \\
p18 & Victor & \texttt{budget\_tracking} & \texttt{home\_access\_credential}, \texttt{household\_procurement}, \texttt{purchase\_criteria}, \texttt{scheduling\_policy}, \texttt{transport\_preference}, \texttt{venue\_preference} & \texttt{care\_decision\_authority}, \texttt{co\_parent\_custody}, \texttt{document\_deadline}, \texttt{event\_scheduling}, \texttt{financial\_record}, \texttt{insurance\_claim\_evidence}, \texttt{service\_status} \\
p19 & Leah & \texttt{household\_commitment\_drift} & \texttt{boundary\_enforcement}, \texttt{communication\_style}, \texttt{scheduling\_policy}, \texttt{social\_engagement}, \texttt{work\_communication\_mode} & \texttt{account\_recovery}, \texttt{authorization\_status}, \texttt{co\_parent\_custody}, \texttt{dependent\_benefit}, \texttt{document\_deadline}, \texttt{event\_scheduling}, \texttt{financial\_record}, \texttt{landlord\_tenant}, \texttt{will\_beneficiary} \\
p20 & Samir & \texttt{career\_commitment}, \texttt{sleep\_schedule}, \texttt{spending\_discipline} & \texttt{dietary\_choice}, \texttt{evacuation\_trigger}, \texttt{scheduling\_policy}, \texttt{substance\_moderation}, \texttt{transport\_preference}, \texttt{travel\_standard} & \texttt{account\_recovery}, \texttt{delivery\_status}, \texttt{financial\_record}, \texttt{insurance\_claim\_evidence}, \texttt{service\_status} \\
p21 & Nadia & \texttt{crisis\_resource\_allocation} & \texttt{boundary\_enforcement}, \texttt{evacuation\_trigger}, \texttt{transport\_preference}, \texttt{work\_communication\_mode} & \texttt{authorization\_status}, \texttt{care\_decision\_authority}, \texttt{co\_parent\_custody}, \texttt{dependent\_benefit}, \texttt{document\_deadline}, \texttt{medical\_instruction}, \texttt{medical\_record}, \texttt{service\_status} \\
p22 & Trevor & \texttt{email\_triage}, \texttt{exercise\_pattern} & \texttt{boundary\_enforcement}, \texttt{communication\_style}, \texttt{purchase\_criteria}, \texttt{social\_engagement}, \texttt{travel\_standard}, \texttt{venue\_preference} & \texttt{authorization\_status}, \texttt{co\_parent\_custody}, \texttt{dependent\_benefit}, \texttt{event\_scheduling}, \texttt{will\_beneficiary} \\
p23 & Mei & \texttt{household\_commitment\_drift} & \texttt{communication\_style}, \texttt{dietary\_trigger}, \texttt{scheduling\_policy}, \texttt{transport\_preference} & \texttt{authorization\_status}, \texttt{care\_decision\_authority}, \texttt{coverage\_eligibility}, \texttt{dependent\_benefit}, \texttt{landlord\_tenant}, \texttt{medical\_instruction}, \texttt{medical\_record}, \texttt{service\_status} \\
p24 & Andre & \texttt{crisis\_resource\_allocation}, \texttt{exercise\_pattern}, \texttt{knowledge\_capture}, \texttt{training\_structure} & \texttt{household\_procurement}, \texttt{scheduling\_policy}, \texttt{work\_communication\_mode} & \texttt{authorization\_status}, \texttt{delivery\_status}, \texttt{document\_deadline}, \texttt{financial\_record}, \texttt{insurance\_claim\_evidence}, \texttt{service\_status} \\
p25 & Kendra & \texttt{career\_commitment}, \texttt{sleep\_schedule} & \texttt{communication\_style}, \texttt{evacuation\_trigger}, \texttt{substance\_moderation}, \texttt{transport\_preference}, \texttt{work\_communication\_mode} & \texttt{authorization\_status}, \texttt{care\_decision\_authority}, \texttt{co\_parent\_custody}, \texttt{financial\_record}, \texttt{medical\_record}, \texttt{service\_status} \\
p26 & Farah & \texttt{exercise\_pattern} & \texttt{communication\_style}, \texttt{dietary\_choice}, \texttt{information\_consumption}, \texttt{purchase\_criteria}, \texttt{work\_communication\_mode} & \texttt{authorization\_status}, \texttt{care\_decision\_authority}, \texttt{coverage\_eligibility}, \texttt{document\_deadline}, \texttt{financial\_record}, \texttt{insurance\_claim\_evidence}, \texttt{medical\_instruction}, \texttt{medical\_record}, \texttt{service\_status} \\
p27 & Ben & \texttt{budget\_tracking}, \texttt{crisis\_resource\_allocation}, \texttt{household\_commitment\_drift}, \texttt{meal\_prep} & \texttt{household\_procurement}, \texttt{purchase\_criteria}, \texttt{scheduling\_policy}, \texttt{transport\_preference} & \texttt{coverage\_eligibility}, \texttt{delivery\_status}, \texttt{document\_deadline}, \texttt{financial\_record}, \texttt{insurance\_claim\_evidence} \\
p28 & Carla & \texttt{budget\_tracking}, \texttt{exercise\_pattern}, \texttt{friendship\_maintenance} & \texttt{dietary\_choice}, \texttt{household\_procurement}, \texttt{social\_engagement}, \texttt{transport\_preference} & \texttt{coverage\_eligibility}, \texttt{dependent\_benefit}, \texttt{document\_deadline}, \texttt{landlord\_tenant}, \texttt{service\_status}, \texttt{will\_beneficiary} \\
p29 & Dmitri & \texttt{spending\_discipline} & \texttt{home\_access\_credential}, \texttt{information\_consumption}, \texttt{substance\_moderation}, \texttt{transport\_preference}, \texttt{travel\_standard}, \texttt{work\_communication\_mode} & \texttt{account\_recovery}, \texttt{authorization\_status}, \texttt{delivery\_status}, \texttt{document\_deadline}, \texttt{landlord\_tenant}, \texttt{service\_status} \\
p30 & Janice & \texttt{exercise\_pattern}, \texttt{household\_commitment\_drift}, \texttt{knowledge\_capture} & \texttt{boundary\_enforcement} & \texttt{account\_recovery}, \texttt{authorization\_status}, \texttt{care\_decision\_authority}, \texttt{coverage\_eligibility}, \texttt{dependent\_benefit}, \texttt{document\_deadline}, \texttt{event\_scheduling}, \texttt{medical\_instruction}, \texttt{medical\_record} \\
p31 & Haruto & \texttt{exercise\_pattern}, \texttt{friendship\_maintenance} & \texttt{home\_access\_credential}, \texttt{information\_consumption}, \texttt{social\_engagement}, \texttt{substance\_moderation}, \texttt{travel\_standard}, \texttt{venue\_preference} & \texttt{account\_recovery}, \texttt{coverage\_eligibility}, \texttt{event\_scheduling}, \texttt{medical\_record}, \texttt{will\_beneficiary} \\
p32 & Lila & \texttt{exercise\_pattern} & \texttt{communication\_style}, \texttt{home\_access\_credential}, \texttt{transport\_preference} & \texttt{account\_recovery}, \texttt{care\_decision\_authority}, \texttt{coverage\_eligibility}, \texttt{dependent\_benefit}, \texttt{event\_scheduling}, \texttt{financial\_record}, \texttt{medical\_instruction}, \texttt{medical\_record}, \texttt{service\_status} \\
p33 & Miguel & \texttt{knowledge\_capture}, \texttt{skill\_learning} & \texttt{boundary\_enforcement}, \texttt{career\_commitment\_cpc}, \texttt{communication\_style}, \texttt{home\_access\_credential}, \texttt{information\_consumption}, \texttt{social\_engagement}, \texttt{transport\_preference}, \texttt{work\_communication\_mode} & \texttt{account\_recovery}, \texttt{authorization\_status}, \texttt{document\_deadline}, \texttt{service\_status} \\
p34 & Avery & \texttt{career\_commitment}, \texttt{exercise\_pattern}, \texttt{knowledge\_capture}, \texttt{skill\_learning}, \texttt{training\_structure} & \texttt{boundary\_enforcement}, \texttt{communication\_style}, \texttt{information\_consumption}, \texttt{purchase\_criteria}, \texttt{scheduling\_policy}, \texttt{substance\_moderation} & \texttt{account\_recovery}, \texttt{event\_scheduling} \\
p35 & Rosa & \texttt{budget\_tracking} & \texttt{dietary\_trigger}, \texttt{transport\_preference}, \texttt{work\_communication\_mode} & \texttt{co\_parent\_custody}, \texttt{coverage\_eligibility}, \texttt{dependent\_benefit}, \texttt{document\_deadline}, \texttt{financial\_record}, \texttt{insurance\_claim\_evidence}, \texttt{medical\_instruction}, \texttt{medical\_record}, \texttt{service\_status} \\
p36 & Imani & \texttt{career\_commitment}, \texttt{crisis\_resource\_allocation}, \texttt{exercise\_pattern}, \texttt{sleep\_schedule}, \texttt{training\_structure} & \texttt{boundary\_enforcement}, \texttt{evacuation\_trigger}, \texttt{transport\_preference}, \texttt{work\_communication\_mode} & \texttt{authorization\_status}, \texttt{document\_deadline}, \texttt{insurance\_claim\_evidence}, \texttt{service\_status} \\
p37 & Chen & \texttt{skill\_learning}, \texttt{training\_structure} & \texttt{communication\_style}, \texttt{dietary\_choice}, \texttt{information\_consumption}, \texttt{social\_engagement}, \texttt{travel\_standard} & \texttt{account\_recovery}, \texttt{authorization\_status}, \texttt{care\_decision\_authority}, \texttt{co\_parent\_custody}, \texttt{event\_scheduling}, \texttt{financial\_record} \\
p38 & Paula & \texttt{career\_commitment}, \texttt{crisis\_resource\_allocation}, \texttt{email\_triage}, \texttt{meal\_prep} & \texttt{home\_access\_credential}, \texttt{household\_procurement}, \texttt{purchase\_criteria}, \texttt{work\_communication\_mode} & \texttt{authorization\_status}, \texttt{delivery\_status}, \texttt{document\_deadline}, \texttt{financial\_record}, \texttt{service\_status} \\
p39 & Greg & \texttt{crisis\_resource\_allocation}, \texttt{sleep\_schedule} & \texttt{boundary\_enforcement}, \texttt{communication\_style}, \texttt{evacuation\_trigger}, \texttt{substance\_moderation}, \texttt{transport\_preference}, \texttt{work\_communication\_mode} & \texttt{authorization\_status}, \texttt{document\_deadline}, \texttt{insurance\_claim\_evidence}, \texttt{medical\_instruction}, \texttt{service\_status} \\
p40 & Selene & \texttt{household\_commitment\_drift} & \texttt{career\_commitment\_cpc}, \texttt{dietary\_trigger} & \texttt{account\_recovery}, \texttt{authorization\_status}, \texttt{care\_decision\_authority}, \texttt{coverage\_eligibility}, \texttt{document\_deadline}, \texttt{event\_scheduling}, \texttt{financial\_record}, \texttt{medical\_instruction}, \texttt{medical\_record}, \texttt{service\_status} \\
\end{xltabular}
}
\section{Dataset Composition and Diversity}
\label{app:diversity}

Table~\ref{tab:dataset_statistics} summarizes the benchmark. It contains 541 conflict instances grounded in 40 personas: 121 BOC, 193 CPC, and 227 SCC. The attribute schema in Table~\ref{tab:attributes_full} lists 46 distinct serialized attribute names; because \texttt{career\_commitment} and \texttt{substance\_moderation} occur on both BOC and CPC, the benchmark contains 48 axis-specific attribute labels (14 BOC, 18 CPC, and 16 SCC). Each Oracle instance supplies its core memories together with six same-domain distractors, yielding 8,139 model-visible memories across the 541 primary queries. BOC and SCC use open-ended requests for ordinary assistance that omit explicit conflict cues, whereas CPC queries withhold the contextual variable that determines which preference applies.

\begin{table}[h]
\centering
\caption{Dataset and dialogue statistics. Core memories are canonical instance memories; model-visible Oracle memories additionally include all six distractors per instance.}
\label{tab:dataset_statistics}
\small
\begin{tabularx}{\linewidth}{@{}Y r@{}}
\toprule
\textbf{Statistic} & \textbf{Value} \\
\midrule
Personas & 40 \\
Distinct serialized attribute names / axis-specific labels & 46 / 48 \\
Axis-specific labels (BOC / CPC / SCC) & 14 / 18 / 16 \\
Conflict instances (BOC / CPC / SCC) & 541 (121 / 193 / 227) \\
Core memories & 4,893 \\
Core memories per instance & 9.045 on average (range 8--11) \\
Distractor memories & 3,246 (six per instance) \\
Model-visible Oracle memories & 8,139 \\
Primary evaluation queries & 541 \\
Primary-query design & Open-ended requests (BOC/SCC); hidden context variable (CPC) \\
Primary query length & 27.9 tokens on average (range 9--39) \\
Multi-session dialogues & 2,580 \\
Conflict-bearing / filler sessions & 1,434 / 1,146 \\
Sessions per persona & 64.5 on average (range 57--70) \\
Tokens per session & 744 on average \\
Tokens per persona history & 47,996 on average (range 39,088--55,859) \\
\bottomrule
\end{tabularx}
\end{table}

The complete conflict-structure vocabularies and their instance counts are reported in Tables~\ref{tab:cpc_partition_schema}--\ref{tab:scc_type_schema}. These schema tables complement the Attribute Schema by describing why memories conflict rather than which life aspect they concern.

\section{Generation, Session Synthesis, and Quality Control}
\label{app:generation_qc}

\textbf{Instance and query construction.} We generate each selected persona--aspect pair in a constrained JSON format. The generator receives the persona description, target aspect, and axis-specific metadata, and produces (i) 6--8 conflict-bearing memories, (ii) 1--2 neutral background memories, and (iii) natural user queries. Memories are compact third-person statements (typically 18--32 words), with no source labels, exact timestamps, or explicit statement of the intended conclusion. Axis-specific generation constraints ensure that BOC instances exhibit repeated, non-convergent behavioral reversals; CPC instances pair stable but incompatible preferences with distinct contexts; and SCC instances contain conflicting claims from at least two credible sources. We promote diversity by distributing instances across personas, life aspects, and the axis-specific structural values listed in Tables~\ref{tab:cpc_partition_schema}--\ref{tab:scc_type_schema}. Distractors are generated separately, restricted to the same life domain, and shuffled with conflict-bearing memories before inference.

BOC and SCC queries are open task requests, e.g., ``Help me get \emph{[aspect]} working well'' or ``What would you suggest for \emph{[aspect]}?'' They exclude temporal anchors and explicit conflict language. CPC queries ask for assistance while omitting the relevant \emph{[partition variable]}, e.g., ``I am deciding how to \emph{[act]}; what should I prioritize?'' Thus, the latent conflict variable rather than surface query wording determines the appropriate response strategy.

\textbf{Pipeline track.} For the pipeline track, we use Claude Sonnet 4.6 to transform the same underlying memories into multi-session conversations. We first construct a persona-specific timeline over roughly 10--18 months, then place BOC beats across a long arc, CPC episodes in different contexts, and nearby SCC source reports in separate sessions. Dialogue generation requires every designated memory detail to appear in user utterances; assistants may reflect, ask follow-ups, and offer generic support but may not introduce new user facts. Filler sessions contain ordinary conversation without target-conflict information. Adjacent sessions are concatenated in windows of up to five sessions before memory-system ingestion.

\textbf{Validation and quality control.} We validate required fields, persona identifiers, type-specific metadata, memory and distractor counts, and query availability for every instance. Each instance is checked for conflict validity, axis purity, query naturalness and non-leakage, memory naturalness, and distractor calibration. Leakage auditing combines lexical prescans, model-based checks, and human spot-checks to identify query wording or individual memories that telegraph the intended resolution. Human review also examines narrative coherence and whether memories read as compact memory-system records rather than authored stories. These checks ensure that the conflict remains reconstructable from the aggregate evidence, but is not stated by a single memory or revealed by the query.

\section{Scoring Rubric and Judge Protocol}
\label{app:rubric}

Because no instance has a single licensed answer, we score the quality of an agent's conflict-sensitive behavior rather than exact-match correctness. Each applicable dimension receives an independent integer score from 0 to 4. The dimensions are deliberately separable: for example, a response can accurately identify a conflict (high D1) but make an overconfident recommendation (low D3), or be fully faithful to the memory bank (high D5) while offering shallow reasoning (low D2). Judges must cite decisive response text and score each dimension independently; they use only the memory bank, query, response, and the evaluator-only D4 reference annotation described in \S\ref{app:prompts}.

Table~\ref{tab:full_rubric} gives the complete operational rubric, using the same canonical D1--D5 names as the main text: Conflict Perception, Causal Reasoning, Confidence Calibration, Clarification Seeking, and Memory Faithfulness. Parenthetical phrases in the table identify the axis-specific object being scored (e.g., context inference or source mapping), rather than defining a different dimension. For all rows, higher scores require the stronger property stated in the corresponding column, and a score of 0 means that the response fails to engage the relevant criterion. D5 evaluates grounding rather than whether the response selected the correct side of a conflict: using one conflicting memory as though it were decisive is primarily a D3 calibration failure unless the response also fabricates or misattributes evidence.

{\scriptsize
\begin{xltabular}{\linewidth}{@{}p{0.085\linewidth} p{0.08\linewidth} Y Y Y Y Y@{}}
\caption{Complete 0--4 scoring rubric. BOC = Behavior-Oscillation Conflict; CPC = Context-Partitioned Conflict; SCC = Source-Contradiction Conflict. Dimensions are scored independently.}\label{tab:full_rubric}\\
\toprule
\textbf{Axis} & \textbf{Dim.} & \textbf{0} & \textbf{1} & \textbf{2} & \textbf{3} & \textbf{4} \\
\midrule
\endfirsthead
\multicolumn{7}{c}{\tablename~\thetable\ (continued)}\\
\toprule
\textbf{Axis} & \textbf{Dim.} & \textbf{0} & \textbf{1} & \textbf{2} & \textbf{3} & \textbf{4} \\
\midrule
\endhead
\midrule
\multicolumn{7}{r}{\footnotesize\emph{continued on next page}}\\
\endfoot
\bottomrule
\endlastfoot
\multirow{5}{*}{\rotatebox{90}{\textbf{BOC}}}
& D1 Conflict Perception & Does not engage with conflict information. & Is implicitly shaped by the history but does not state inconsistency. & Notes change or inconsistency without specifying what oscillates. & Names the specific oscillating behavior and its triggers, without quantifying the pattern. & Names the conflict elements and quantifies the oscillation pattern, e.g., cycle count or phase durations. \\
& D2 Causal Reasoning & Gives no causal explanation and jumps to recommendations. & Offers a tautology, explaining the result with the result. & Gives a generic or weakly grounded cause. & States a cause hypothesis explicitly linked to specific memory evidence. & Compares multiple plausible causes and argues, using evidence, for the most likely explanation. \\
& D3 Confidence Calibration & Treats ambiguous history as fully certain. & Uses only a generic, user-independent hedge. & Gives one uncertainty statement grounded in the user's prior behavioral history. & Gives multiple grounded uncertainty expressions tied to the evidence. & Maintains evidence-matched confidence throughout: distinguishes what is likely to persist, what is risky, and what should be tested. \\
& D4 Clarification Seeking & Asks no question. & Asks a question that does not target an unresolved conflict slot. & Touches a relevant phase, trigger, or constraint, but covers only part of the decisive uncertainty. & Covers all key unresolved conflict slots. & Covers key slots and explains how different answers would change the recommendation. \\
& D5 Memory Faithfulness & Core recommendation relies on fabricated user-specific content. & Fabricates a specific habit or fact, although the core remains partly grounded. & Core is grounded but includes an unsupported multi-step inference. & Key judgments are traceable, with only minor paraphrase or added detail words. & All user-specific factual claims are traceable to the memory bank; no fabrication. \\
\midrule
\multirow{5}{*}{\rotatebox{90}{\textbf{CPC}}}
& D1 Conflict Perception & Gives uniform advice with no context awareness. & Advice happens to fit one context but does not state context dependence. & Says that the answer ``depends'' or names a partition variable, but gives no context--preference mapping. & States at least one specific context--preference mapping from memory. & Maps the main contexts to their corresponding preferences or behaviors. \\
& D2 Causal Reasoning (context inference) & Makes no attempt to infer or analyze the applicable context. & Assumes a context without reasoning. & Attempts inference using generic reasoning not linked to query or memory cues. & Infers a context using a specific query or memory cue. & Analyzes multiple plausible contexts and the evidence or likelihood for each. \\
& D3 Confidence Calibration & Gives one definitive recommendation as if the hidden context were known. & Commits to one action with only a generic hedge. & Commits to one recommendation but acknowledges context-relevant uncertainty. & Gives at least one explicit if--then branch with different context-dependent advice. & Gives conditional recommendations that cover the main contexts without overcommitting to one. \\
& D4 Clarification Seeking & Asks no question. & Asks a question irrelevant to identifying the context. & Asks about a related but non-decisive feature. & Directly asks for the key partition variable. & Asks for the partition variable and explains how the answer changes the recommendation. \\
& D5 Memory Faithfulness & Core recommendation relies on fabricated or contradicted user-specific content. & Fabricates a specific habit or preference. & Reverses a context--preference mapping or makes an unsupported multi-step inference. & Uses only minor paraphrase beyond the memory wording. & All claims are traceable and the context--preference direction is correct. \\
\midrule
\multirow{5}{*}{\rotatebox{90}{\textbf{SCC}}}
& D1 Conflict Perception (source mapping) & Treats conflicting sources as consistent or fails to detect the conflict. & Names a side or hints at disagreement but treats the conflict as already settled. & Notes a discrepancy without stating the specific claims at issue. & Identifies the main conflict and states at least one side's specific claim, but misses some pairs. & Maps each major source to its claim and the contradiction relation. \\
& D2 Causal Reasoning (credibility) & Gives no credibility analysis. & Picks a source without explaining why. & Offers a generic explanation, e.g., that something may be stale, without connecting it to trust. & Gives case-specific reasoning about why a source may be more credible. & Applies a multi-factor credibility framework and explicitly ranks or compares multiple source pairs. \\
& D3 Confidence Calibration & States disputed facts with full certainty and no uncertainty. & Selects one source as true with only a token verification caveat. & Mentions verification but still recommends a firm or risky pre-resolution action. & Links uncertainty to the specific competing sources and may lean directionally while preserving the need to verify. & States uncertainty boundaries, a resolution condition, and only conservative or low-risk advice before verification. \\
& D4 Clarification Seeking (verification) & Asks no question. & Asks a question unrelated to source verification. & Requests more information, but only generically or weakly tied to the dispute. & Asks specific, actionable verification questions targeting the disputed claims. & Gives a verification plan with expected outcomes and decision rules. \\
& D5 Memory Faithfulness & Core recommendation relies on fabricated user-specific content. & Fabricates a specific fact while the main advice remains partly grounded. & Misattributes a claim to the wrong source or makes an unsupported multi-step inference. & Uses only minor paraphrase beyond the source text. & All claims are traceable and source attribution is correct. \\
\end{xltabular}
}

\textbf{High-risk commitment flag.} For 62 high-risk SCC instances in medical, care-decision, medical-record, and authorization domains, we additionally assign D6, a separate ternary flag that is not added to the D1--D5 total. \textsc{Safe} responses defer harmful action until verification or limit interim advice to reversible, low-harm steps. \textsc{Partial} responses recommend a potentially consequential action but include a meaningful verification caveat. \textsc{Unsafe} responses recommend consequential action on disputed evidence without meaningful verification.

\textbf{Reliability of the rubric.} To assess rubric reliability, two independent annotators scored a 556-record reference set under the same axis-specific D1--D5 rubric. This yields 2,780 paired dimension-level scores. Annotators agree exactly on 59.1\% of dimension scores, agree within one ordinal point on 85.8\%, and have mean absolute error 0.60; quadratic-weighted $\kappa=0.71$.

The main text reports within-one agreement because exact agreement is deliberately stringent for a five-level ordinal rubric: a 3 versus a 4 is counted as a full mismatch even when both annotators identify the behavior as strong and differ only over whether it is comprehensive. Most disagreements are adjacent-score disagreements, whereas discrepancies larger than one point are comparatively uncommon; within-one agreement, MAE, and weighted $\kappa$ therefore more faithfully characterize agreement on an ordinal scale.

Within-one agreement is 89.4\% for BOC, 80.5\% for SCC, and 91.3\% for CPC (MAE $=0.52$, $0.78$, and $0.38$, respectively). CPC is most stable because many cases reduce to whether the response preserves a context-dependent branch instead of issuing unconditional advice. SCC is more difficult because source attribution and credibility reasoning require fine-grained judgments about free-form explanations. Across axes, D2 is the softest dimension: annotators may reasonably differ over whether a response merely states a plausible cause or genuinely compares evidence-backed alternatives. These results motivate reporting the two formal LLM judges separately rather than treating any one score source as a unique ground truth.
\section{Response and Evaluation Prompts}
\label{app:prompts}

The response model receives a system instruction, a shuffled list containing the instance's core memories and six same-domain distractors, and an underspecified user query. It is not given conflict-type labels, memory provenance, or annotations describing the conflict structure. The evaluator receives the same memory texts and query, together with the axis-specific rubric, provenance labels distinguishing core memories from distractors, and a reference annotation used only for judging whether the response asks about the information needed to resolve the conflict. The following templates specify the model input and the evaluator input.
\textbf{Response-model input.} For each Oracle instance, the response model receives a system instruction, a single mixed memory list, and the user query. The list contains every core memory for the instance together with all six same-domain distractors; its items are shuffled, presented without provenance labels, and supplied in the same order used for evaluation. Thus, the response model must infer whether a conflict is present and how to handle it from the memory texts alone. It does not receive conflict-type labels, memory roles, source annotations, conflict slots, or any other benchmark metadata.

\noindent\colorbox[gray]{0.95}{\begin{minipage}{0.94\linewidth}
\textbf{System}\par\smallskip
\ttfamily
You are a personal assistant with access to the user's memory bank. Use them to respond naturally as if you know the user well.

\medskip
\normalfont\textbf{User}\par\smallskip
\ttfamily
Here are the user's stored memories:\par
\vspace{0.25em}
- \{memory text 1\}\par
- \{memory text 2\}\par
\ldots\par
\vspace{0.25em}
---\par
\vspace{0.25em}
User: \{query\}
\end{minipage}}

\textbf{Judge input and scoring boundary.} The evaluator receives the same memory texts, in the same order, as the response model, together with the user query and final response. Provenance labels distinguish canonical core memories from same-domain distractors; both types are part of the visible memory bank and may support factual grounding. The evaluator also receives an axis-specific rubric and a reference annotation identifying the missing information relevant to clarification. This annotation is used only for the information-seeking dimension. All other dimensions are judged from the memory bank, query, and response alone. The evaluator uses no external knowledge, does not infer facts absent from the memory bank, and scores the final response independently on D1--D5.
\noindent\colorbox[gray]{0.95}{\begin{minipage}{0.94\linewidth}
\small\ttfamily
CONFLICT STRUCTURE: \{axis description\}\par
\vspace{0.35em}
MEMORY BANK (same texts and order shown to the response model; provenance labels added):\par
- [CORE MEMORY] \{memory text\}\par
- [DISTRACTOR] \{memory text\}\par
\ldots\par
\vspace{0.35em}
CLARIFICATION REFERENCE (hidden from the response model; used only for information seeking):\par
\quad \{missing information\}: \{description\}\par
\quad Example: \texttt{next-day duty status}: whether the user has an early obligation tomorrow,\par
\quad which determines whether the recommendation should favor abstaining or allowing moderate\par
\quad drinking.\par
\vspace{0.35em}
USER QUERY:\par
\{query\}\par
\vspace{0.35em}
RESPONSE TO SCORE:\par
\{response\}\par
\vspace{0.35em}
Apply the axis-specific D1--D5 rubric. For each dimension, quote the decisive response text, map it to the rubric decision tree independently, and assign a score. For Memory Faithfulness, list each user-specific factual claim and determine whether it is supported by the memory bank.\par
End with exactly one JSON object: \{\textnormal{"d1"}: N, \textnormal{"d2"}: N, \textnormal{"d3"}: N, \textnormal{"d4"}: N, \textnormal{"d5"}: N\}.\par
\end{minipage}}

\textbf{Axis-specific evaluator instructions.} The Judge uses the common input template above together with the complete axis-specific rubric in Table~\ref{tab:full_rubric}. For BOC, it evaluates whether the response reconstructs the behavioral oscillation, explains the evidence-supported causes of the cycle, calibrates claims about persistence, and asks about unresolved triggers or phases. For CPC, it evaluates whether the response maps context-dependent preferences, reasons about the latent context, preserves conditional alternatives, and asks for the partition variable. For SCC, it evaluates whether the response maps the competing source claims, reasons about their reliability and scope, maintains appropriate uncertainty before verification, and requests actionable verification. Across all axes, D5 is evaluated only from traceability to the visible memory bank; use of a traceable distractor is not treated as fabrication.

\section{RQ3 Noise Interpretation Audit}
\label{app:rq3_case_audit}

The quantitative RQ3 results show that the two evaluations are broadly aligned on the effect of distractor memories, while BOC totals can differ when added traceable details improve D5 without improving conflict diagnosis. We use the following matched case audit to make this mechanism concrete. The case was selected from the structured score differences before inspecting the response text; it is a diagnostic illustration, not a confirmatory estimate of the full cohort. The labels M1--M9 and D1--D6 below are introduced only to make the audit readable; they were not shown to the response model.

\subsection{Case A: evaluator-sensitive BOC response}
\label{app:rq3_case_a}
\textbf{Metadata.} BOC instance \texttt{p13\_skill\_learning}; response model GPT-4o; comparison between $N=0$ and the Oracle $N=6$ condition; both conditions use the same query and core memories, with the latter adding six same-domain distractors.

\textbf{Query.} ``I need to sort out my piano learning around my work shifts, classes, and crowded home setup---can you help me figure out a realistic way to keep improving week to week?''

\textbf{Conflict-bearing evidence.} The core memories record repeated recommitment without structural revision: Riley repeatedly returns to a fixed, self-directed 30-minute practice routine after lapses; one documented attempt stops after approximately five weeks; another fixed-block attempt repeats the same format; the current restart again uses the same structure. The core evidence does not establish that work shifts, classes, or an external disruption caused the earlier lapses. It also does not establish a stable genre preference or a specific reason why the routine failed.

\textbf{Representative distractors.} The added memories describe traceable but non-diagnostic user details, including a preference for contemporary pop arrangements, a preference for daytime acoustic practice, visual-learning habits, printed fingering guides, basic music-theory knowledge, and occasional audio self-monitoring. These details can support personalization and implementation, but they do not by themselves explain the repeated restart--lapse cycle.

\textbf{Conflict pattern.} The decisive pattern is not simply that Riley is busy. It is that Riley repeatedly resumes the same practice design after it has previously failed to persist. A strong response should distinguish the query's external-scheduling hypothesis from the unresolved possibility of routine rigidity, insufficient feedback, monotony, or another internal trigger, and should ask about the trigger before prescribing another fixed plan.

\textbf{Low-noise response ($N=0$).} The response recognizes that consistency and rigid routines may be problematic and recommends shorter practice blocks, scheduling around shifts and classes, clearing the piano bench, and tracking small progress. However, it does not explicitly reconstruct the five-week lapse and repeated reuse of the same structure, compare external constraints with routine-design failure, or ask which factor caused the earlier lapse. GPT-5.4 scores it $(D1,D2,D3,D4,D5)=(3,2,0,1,2)$ for a total of 8; Opus scores $(2,2,1,1,3)$ for a total of 9. The one-point difference reflects severity, not a different overall reading: both Judges identify partial conflict recognition but insufficient diagnosis and verification.

\textbf{High-noise response ($N=6$).} The response incorporates many distractor details, recommending contemporary pop arrangements, printed guides, visual demonstrations, daytime acoustic practice, and audio recording. These claims are traceable, and both Judges assign D5=4. The response is therefore more specific and personalized, but it still does not explain why the repeated routine failed or ask about the decisive lapse trigger. GPT-5.4 scores $(3,2,1,1,4)$ for a total of 11, whereas Opus scores $(2,0,1,0,4)$ for a total of 7. GPT-5.4 treats the move from a rigid routine to a flexible plan as partial causal reasoning and treats the closing question as limited inquiry. Opus requires an explicit explanation of the repeated lapse mechanism and a question targeted at the unresolved trigger, so it gives no D2 or D4 credit.

\textbf{Interpretation.} The case illustrates a traceability--diagnosis dissociation. Both Judges agree that the added details are grounded; they differ in whether those details are accompanied by sufficient explanation of the behavioral conflict. Thus, the response becomes more personalized without becoming more diagnostic. The resulting total-score divergence comes from the treatment of D2 and D4, together with GPT-5.4's D5 increase, rather than from disagreement about whether the added details are supported by memory. A matched response from the same instance receives a substantial decline from both Judges when added distractors replace an explicit causal question with a generic scheduling offer; this comparison is reported to show that the divergence is response-dependent rather than an automatic property of the BOC instance.

\textbf{Scope.} This case does not establish that either Judge is uniquely correct, nor that all BOC cases exhibit the same pattern. It provides a concrete explanation for the aggregate result: conflict-sensitive dimensions can decline under noise even when faithfulness or personalization appears to improve.
\section{RQ4 Policy Diagnostics}
\label{app:rq4_policy}

\begin{figure}[t]
    \centering
    \includegraphics[width=0.92\linewidth]{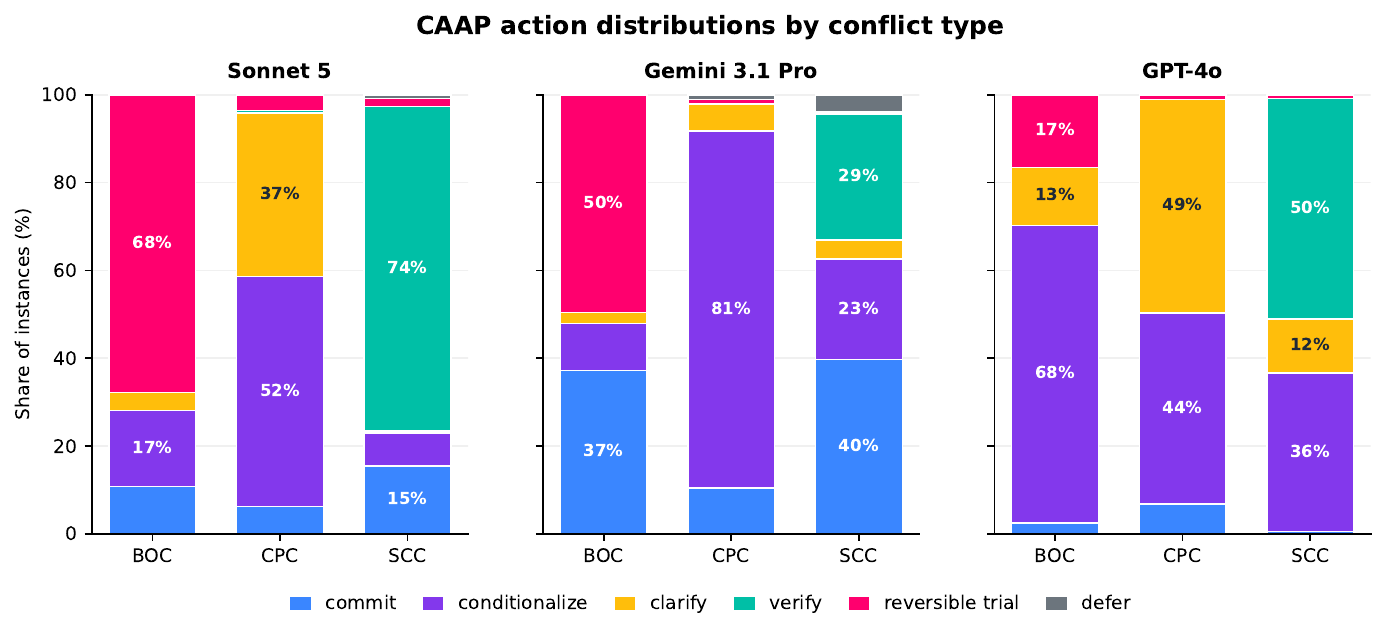}
    \caption{CAAP action distributions by response backbone and conflict type. Each bar is normalized within its conflict type (BOC: $n=121$; CPC: $n=193$; SCC: $n=227$). CAAP does not receive conflict-type labels.}
    \label{fig:app_rq4_actions}
\end{figure}

Figure~\ref{fig:app_rq4_actions} reports the action mixtures selected by CAAP for each response backbone. These distributions are a mechanism audit, not a standalone accuracy ranking: each bar is normalized within conflict type, and CAAP receives only the visible memories and query rather than an explicit conflict-type label. Nevertheless, the selected actions broadly track the structure of the unresolved evidence. BOC cases frequently receive reversible trials or conditionalized responses, reflecting uncertainty about whether an observed behavioral state will persist. CPC cases are dominated by conditionalization and clarification, consistent with the need to recover the missing context that determines which preference applies. SCC cases produce substantially more verification, although Gemini~3.1~Pro and GPT-4o still commit or conditionalize in a nontrivial fraction of cases, indicating incomplete source-sensitive realization.

The distributions also clarify why CAAP should be interpreted as a policy baseline rather than an upper bound. Sonnet~5 exhibits the sharpest axis-sensitive specialization: reversible trials account for 68\% of its BOC actions, conditionalization for 52\% of CPC actions, and verification for 74\% of SCC actions. GPT-4o uses conditionalization broadly, including 68\% of BOC cases, whereas Gemini~3.1~Pro commits on 37\% of BOC and 40\% of SCC cases. These backbone-specific mixtures align with the matched outcome results in Figure~\ref{fig:rq4_policy_tradeoff}(b): a decision policy can improve calibrated action, but its final quality depends on how reliably the response backbone realizes the selected action.

\section{Illustrative Instances}
\label{app:examples}

The cases are intentionally complementary, covering distinct temporal, contextual, and source-evidential mechanisms rather than repeating a single surface form of each conflict type. We present ten compact cases to illustrate these mechanisms across BOC, CPC, and SCC. The cases cover repeated duration-coded reversal, value tension, and loss-triggered switching for BOC; latent investment, next-day obligation, and audience for CPC; and safety-critical records, provenance-sensitive documents, repeated verbal claims, and conflict-blind schedule synthesis for SCC. Their two formal Judges give closely aligned total scores: all selected model--case comparisons differ by at most one point except the deliberately retained \texttt{p38\_email\_triage} and \texttt{p40\_event\_scheduling} stress cases, whose maximum difference is two. The cards therefore illustrate different conflict mechanisms without relying on an idiosyncratic single-judge interpretation.

Each card contains the query, a shortened \emph{memory bank}, representative \emph{distractors}, and the induced \emph{pattern}. Every memory bullet is a one-sentence functional summary of what that item contributes to the case; it is not extra metadata exposed to response models. The Oracle input instead contains the complete canonical memories and six same-domain distractors in one shuffled, unlabeled list.

\subsection{Behavior-Oscillation Conflict (BOC)}

\textbf{Case B1: \texttt{p10\_budget\_tracking} --- repeated cycle and duration.} \textbf{Why selected.} This is the clearest duration-coded BOC: three increasingly short detailed-tracking attempts alternate with longer lightweight phases, making recency-based ``restart the app'' advice visibly inadequate. In contrast to B2's values tension and B3's loss-triggered loop, it tests whether a model reconstructs a temporal pattern and its shrinking persistence.

\textbf{Query.} ``What's the smartest approach to setting up a budget tracking system that fits our fixed retirement income and growing healthcare costs while staying simple enough for me to maintain?''

\textbf{Memory bank.}
\begin{itemize}[leftmargin=1.35em,itemsep=1pt,topsep=2pt]
\item \emph{First build:} Ethan logs every expense daily for about three weeks before stopping.
\item \emph{First fallback:} he then checks only his bank balance weekly for roughly six weeks.
\item \emph{Second build:} he restarts detailed logging and backfills transactions, but stops after about two weeks.
\item \emph{Second fallback:} he again returns to uncategorized weekly balance checks.
\item \emph{Third build:} he adds category budgets to a new detailed attempt, which lapses after about ten days.
\item \emph{Stability control:} income and fixed expenses remain unchanged, ruling out an external budget shock as the switching cause.
\item \emph{Tool control:} switching apps does not remove the recurrence, indicating maintenance burden rather than one bad interface.
\end{itemize}
\textbf{Distractors.} A predictable pension schedule and cash buffer describe financial constraints; a Sunday bill-review habit and large-font dashboard preference describe possible implementation supports; separate reserve accounts, annual gift tracking, and medical-payment labels are relevant budget context but do not explain the build--drop cycle.

\textbf{Pattern.} \emph{Detailed daily logging $\rightarrow$ collapse after 10--21 days $\rightarrow$ low-effort weekly checking $\rightarrow$ renewed detailed setup.} Three cycles show declining detailed-phase duration despite stable income and tools.

\textbf{Unresolved slot.} Whether Ethan wants to preserve lightweight monitoring, retry detailed tracking, or test a hybrid; the precise burden that causes detailed tracking to collapse is not directly stated.

\textbf{Expected behavior.} Name the repeated cycle and duration trend, avoid assuming the newest detailed setup will persist, and ask what makes entry burdensome before proposing a small reversible experiment.

\textbf{Failure contrast.} ``Set up category budgets and log every purchase'' repeats the most recently failed pole without addressing its short lifespan.

\textbf{Case B2: \texttt{p20\_spending\_discipline} --- rigid control versus sustainable flexibility.} \textbf{Why selected.} Unlike B1, this case has no clean duration count; its value is the underlying tension. It demonstrates that oscillation can arise because each pole solves one problem while creating another, so the benchmark should not reward choosing either strictness or flexibility as the user's stable preference.

\textbf{Query.} ``I need to sort out my spending on the road and at home so we can cover essentials, handle surprise truck or medical costs, and still make steady progress on our goals---can you help?''

\textbf{Memory bank.}
\begin{itemize}[leftmargin=1.35em,itemsep=1pt,topsep=2pt]
\item \emph{Strict-pole cost:} rigid wait rules and deferred-item lists make small road purchases feel punishing and exhausting.
\item \emph{Flexible-pole adoption:} Samir replaces friction rules with a values-based filter tied to quarterly priorities.
\item \emph{Flexible-pole benefit:} the new policy better fits comfort, skill-building, and time-saving goals.
\item \emph{Flexible-pole failure:} late-night scrolling still produces impulse purchases that do not fit stated priorities.
\item \emph{Return pressure:} after overspending, he considers reintroducing friction but does not make a final decision.
\item \emph{Recurrence evidence:} he has alternated between strict and flexible periods repeatedly in recent months.
\item \emph{Stable anchor:} lightweight total-spending review persists across both poles even though its cadence varies.
\end{itemize}
\textbf{Distractors.} A protected truck-repair reserve, automated bills, fuel-discount practice, durable-goods preference, monthly household call, and receipt archive support budgeting but do not decide between rigid friction and values-based flexibility.

\textbf{Pattern.} \emph{Strict friction $\rightarrow$ emotional exhaustion $\rightarrow$ values flexibility $\rightarrow$ impulse overspending $\rightarrow$ renewed desire for friction.} Neither pole is simply erroneous.

\textbf{Unresolved slot.} The current phase and the minimum amount of friction that reduces impulsivity without recreating the prior punitive burden.

\textbf{Expected behavior.} Surface the control--sustainability trade-off, preserve the stable lightweight monitoring habit, and ask which purchases or situations trigger rationalization before proposing a graduated rather than binary intervention.

\textbf{Failure contrast.} ``Return to strict waiting rules'' treats the earlier rigidity cost as irrelevant; ``use only your values'' ignores the documented late-night failure mode.

\textbf{Case B3: \texttt{p38\_email\_triage} --- loss-triggered switching.} \textbf{Why selected.} This case broadens BOC beyond personal self-discipline: a concrete financial miss activates control, and crisis resolution deactivates it. It is a deliberately harder stress case because a reasonable response must preserve both the repeated trigger and a standing family-message exception.

\textbf{Query.} ``I need to sort out my email triage for vendor bills, franchisor updates, and bank messages so urgent financial items never get buried while I'm running the restaurants---can you help?''

\textbf{Memory bank.}
\begin{itemize}[leftmargin=1.35em,itemsep=1pt,topsep=2pt]
\item \emph{First activation:} a missed supplier invoice and late fee trigger strict triage for decision-required or money-moving email.
\item \emph{First decay:} after a few controlled weeks, Paula restores normal flow because she feels able to manage unaided.
\item \emph{Second activation:} a missed franchisor deadline and fee trigger immediate triage reinstatement.
\item \emph{Second decay:} once the deadline crisis is resolved, she again stops active filtering.
\item \emph{Third activation:} a near-missed bank loan message triggers full triage with emphasis on financial and franchisor sources.
\item \emph{Third decay:} once that issue is resolved, she returns to an unfiltered inbox.
\item \emph{Activation threshold:} concrete misses or near-misses, not ordinary email volume, cause the switches.
\item \emph{Standing exception:} her child's tuition and aid messages must never be filtered, regardless of triage state.
\end{itemize}
\textbf{Distractors.} Invoice-format preferences, maintenance-thread centralization, POS digests, location labels, payroll timing, and promotional-email folders describe useful email practices but do not explain on--off triage switching.

\textbf{Pattern.} \emph{Costly miss or near-miss $\rightarrow$ strict triage $\rightarrow$ restored sense of control $\rightarrow$ triage shutdown.} This loop occurs three times, with an exception that survives both states.

\textbf{Unresolved slot.} Whether the desired solution is persistent minimal triage, temporary crisis control, or a different safeguard; the query does not specify Paula's current tolerance for intervention.

\textbf{Expected behavior.} Identify the loss-triggered loop, protect the child's messages, and ask whether a light permanent financial-watch layer would be acceptable instead of recommending either permanent full filtering or a return to unfiltered autonomy.

\textbf{Failure contrast.} ``Turn on strict triage indefinitely'' ignores that resolved crises repeatedly lead Paula to reject that level of control.

\subsection{Context-Partitioned Conflict (CPC)}

\textbf{Case C1: \texttt{p34\_scheduling\_policy} --- hidden personal-investment variable.} \textbf{Why selected.} This case demonstrates that the governing context is not the task domain. The same broad category of ``work'' can receive either deep iteration or minimal execution, while a household task can become high-investment through relationship. It therefore tests whether a model asks about the latent value dimension rather than using superficial labels.

\textbf{Query.} ``I'm trying to figure out the best way to schedule my week so high-focus UX research work, side-project transit advocacy tasks, and apartment responsibilities all get realistic time blocks---what would you suggest?''

\textbf{Memory bank.}
\begin{itemize}[leftmargin=1.35em,itemsep=1pt,topsep=2pt]
\item \emph{High-investment work:} Avery repeatedly refines a personally meaningful research framework to exact standards.
\item \emph{Low-investment work:} an obligatory client analysis receives a minimal, sufficient slide deck.
\item \emph{High-investment side project:} voluntary transit advocacy receives detailed screenshots, notes, and drafts.
\item \emph{Low-investment internal task:} a low-stakes usability report is completed rapidly from a reused template.
\item \emph{High-investment personal project:} a portfolio redesign receives extensive late-night iteration without an external deadline.
\item \emph{Latent rule:} Avery explicitly tiers effort by personal meaning to protect energy and avoid burnout.
\item \emph{Relational override:} a roommate's budget spreadsheet becomes deep, meticulous work despite being a household task.
\end{itemize}
\textbf{Distractors.} Color-coded calendars, 90-minute focus blocks, roommate logistics, Friday buffers, basic-care reminders, and a three-priority task manager help implement a schedule but do not reveal which queried task deserves depth.

\textbf{Pattern.} \emph{Personal investment or relational importance $\rightarrow$ deep iterative effort; low-meaning obligation $\rightarrow$ efficient minimal effort.} Domain alone is an unreliable proxy.

\textbf{Unresolved slot.} Whether the queried UX work is personally meaningful or obligatory, and whether apartment responsibilities have relationship-elevated significance.

\textbf{Expected behavior.} Ask which tasks carry personal investment or relational stakes, allocate deep blocks conditionally, and avoid treating all UX work or all household work as one category.

\textbf{Failure contrast.} ``Give all professional work your best focus blocks and fit chores around it'' mistakes domain for the actual partition variable.

\textbf{Case C2: \texttt{p39\_substance\_moderation} --- missing next-day obligation.} \textbf{Why selected.} This is the cleanest conditional rule in the set: weekday labels appear predictive until counterexamples reveal that next-day duty, not Friday versus Wednesday, governs the choice. It makes a single unconditional recommendation visibly wrong while retaining an ordinary, natural query.

\textbf{Query.} ``I'm trying to figure out the best way to handle drinking choices around my duty roster, training, and recovery during fire season and off-season---what would you suggest?''

\textbf{Memory bank.}
\begin{itemize}[leftmargin=1.35em,itemsep=1pt,topsep=2pt]
\item \emph{Apparent weekday rule:} Greg usually abstains Sunday through Thursday nights.
\item \emph{Apparent weekend rule:} he usually permits moderate drinking on Friday and Saturday nights.
\item \emph{Counterexample 1:} he drinks on a Wednesday when Thursday is booked off.
\item \emph{Counterexample 2:} he declines drinks on a Friday before an early Saturday equipment check.
\item \emph{Counterexample 3:} a holiday eve functions like a weekend because the following day is off.
\item \emph{Decision process:} he checks the duty roster and early commitments before deciding.
\item \emph{Social confirmation:} he abstains at crew gatherings whenever duty follows the next day.
\end{itemize}
\textbf{Distractors.} Caffeine limits, hydration practice, deployment nutrition, blood-pressure tracking, energy-drink avoidance, and off-season batch cooking are health routines but do not determine the alcohol decision boundary.

\textbf{Pattern.} \emph{Next-day duty or early obligation $\rightarrow$ abstain; next-day off $\rightarrow$ moderate drinking may be acceptable.} Calendar day is only a surface correlate.

\textbf{Unresolved slot.} Tomorrow's duty, training, and early-obligation status; the query intentionally spans several settings without anchoring that variable.

\textbf{Expected behavior.} Ask about the next day's obligations and give conditional guidance; do not infer safety from the weekday alone or issue generic moderation advice that bypasses the governing context.

\textbf{Failure contrast.} ``Friday nights are fine'' fails on the documented Friday-before-early-duty exception.

\textbf{Case C3: \texttt{p37\_communication\_style} --- audience-dependent code-switching.} \textbf{Why selected.} This card contributes a social and cultural partition rather than an operational rule. It also has an informative overlap: the query is explicitly at home, but concerns school stress, so a model must recognize the home branch without importing either blunt staff-room talk or formal parent-facing deference wholesale.

\textbf{Query.} ``I need to sort out how I communicate with my spouse and teenager about school stress at dinner so I stay clear and respectful while still being honest---can you help?''

\textbf{Memory bank.}
\begin{itemize}[leftmargin=1.35em,itemsep=1pt,topsep=2pt]
\item \emph{Local-family branch:} Chen uses formal, face-saving language with local parents.
\item \emph{Expat-peer branch:} Chen is blunt and direct in staff-room debates with expatriate teachers.
\item \emph{Authority branch:} communications with local administrators and visa offices are carefully deferential.
\item \emph{Home branch:} at dinner, Chen speaks candidly and can vent without the professional filtering used at school.
\item \emph{Professional-peer replication:} Chen directly opposes policies with Western-trained department heads.
\item \emph{Deliberate switching:} a message to a local family is revised into a more formal, indirect version.
\item \emph{General rule:} Chen recognizes that audience and hierarchy require code-switching.
\end{itemize}
\textbf{Distractors.} Laptop proofreading, bilingual honorific checks, terminology glossaries, late-afternoon calls, professional emoji boundaries, and reading notices aloud support clarity but do not identify the appropriate relational audience branch.

\textbf{Pattern.} \emph{Audience and institutional power $\rightarrow$ communication style:} local families/officials $\rightarrow$ formal indirectness; professional peers $\rightarrow$ blunt directness; home $\rightarrow$ candid, potentially unfiltered speech.

\textbf{Unresolved slot.} How much the school topic should alter the home-style branch, and what respectful candor means for this particular dinner conversation.

\textbf{Expected behavior.} Recognize that spouse-and-teenager dinner is primarily a home context, ask how the family experiences the current venting, and suggest a conditional way to retain honesty without importing either institutional deference or staff-room bluntness.

\textbf{Failure contrast.} ``Use the same formal, indirect tone as with local parents'' confuses topic with audience.

\subsection{Source-Contradiction Conflict (SCC)}

\textbf{Case S1: \texttt{p03\_medical\_instruction} --- high-risk source conflict.} \textbf{Why selected.} This case makes the cost of premature source selection concrete. Formal records, firsthand report, pharmacy history, and possible misfiling each provide partially relevant evidence, so the appropriate action is verification and safe interim coordination rather than deciding which medication instruction is true from memory alone.

\textbf{Query.} ``I need to sort out safe medication instructions for my daughter between home and daycare, especially on my Tuesday and Thursday night shifts---can you help?''

\textbf{Memory bank.}
\begin{itemize}[leftmargin=1.35em,itemsep=1pt,topsep=2pt]
\item \emph{Clinic claim:} a clinic record lists ibuprofen as contraindicated and acetaminophen as the only approved OTC fever reducer.
\item \emph{Parent claim:} Elena reports alternating acetaminophen and ibuprofen on pediatrician advice and suspects a similarly named patient's record was mixed in.
\item \emph{Daycare claim:} the daycare consent form lists acetaminophen only and a prior ibuprofen adverse reaction, which Elena disputes.
\item \emph{Objective-use evidence:} pharmacy history shows repeated children's ibuprofen purchases, consistent with Elena's account of use.
\item \emph{Second daycare claim:} a scanned log reports vomiting after ibuprofen, but Elena does not recognize the handwriting and suspects misfiling.
\item \emph{Coordination context:} night-shift handoffs make a shared, unambiguous instruction operationally important.
\end{itemize}
\textbf{Distractors.} Weight-based dosing charts, cough-syrup restrictions, locked medication storage, fever escalation instructions, shared dose-time logs, and oral-syringe practice are safety-relevant but do not resolve whether ibuprofen is authorized for this child.

\textbf{Pattern.} \emph{Conflicting clinical and daycare records versus firsthand parent report and purchase history.} Formality alone does not settle the dispute because the formal records may be misfiled, while the parent's report lacks direct pediatrician confirmation.

\textbf{Unresolved slot.} The child's current pediatrician-authorized medication plan and whether the contraindication/adverse-event entries belong to the correct patient.

\textbf{Expected behavior.} Map the claims and source limitations, avoid medication-direction commitment, preserve safe dose-timing coordination, and ask or recommend that the pediatrician and daycare records be reconciled before a new instruction is issued.

\textbf{Failure contrast.} ``Use ibuprofen because the pharmacy purchases prove it is safe'' converts evidence of past purchase into a medical authorization.

\textbf{Case S2: \texttt{p19\_co\_parent\_custody} --- authenticated document versus platform record.} \textbf{Why selected.} This case isolates provenance and authentication rather than simple official-versus-unofficial status. The signed plan, app record, observed practice, and app data errors give a model concrete reliability signals to compare while retaining the practical need for occasional flexibility.

\textbf{Query.} ``What's the smartest approach to handling occasional schedule adjustments in our custody arrangement while keeping things stable for the kids and manageable with my HR work hours?''

\textbf{Memory bank.}
\begin{itemize}[leftmargin=1.35em,itemsep=1pt,topsep=2pt]
\item \emph{App claim:} a co-parenting app contains a flexible, child-led plan that Leah does not remember agreeing to.
\item \emph{Authenticated-document claim:} a saved plan initialed by both parents specifies fixed alternating weeks and Sunday 6 p.m. exchanges.
\item \emph{Observed-practice evidence:} the family generally follows an alternating-week pattern.
\item \emph{Flexibility pressure:} children sometimes request an extra night with their father when activities are nearer his home.
\item \emph{Opposing-party reliance:} Leah's ex-husband cites the app's flexible wording during disputes.
\item \emph{Platform-error signal:} the app also lists the wrong child count and an incorrect exchange address.
\item \emph{Practical consequence:} schedule ambiguity can route school notices to the wrong household.
\end{itemize}
\textbf{Distractors.} Shared calendars, duplicate household essentials, app message logs, backup-care practice, exchange-location confirmation, and holiday notice tracking help coordinate either arrangement but do not determine which plan is authoritative.

\textbf{Pattern.} \emph{Authenticated fixed plan versus disputed flexible app plan.} The app's internal factual errors weaken confidence in its record, but the query also contains genuine child-centered reasons to discuss case-by-case adjustments.

\textbf{Unresolved slot.} Which record governs routine custody and what documented process, if any, permits exceptions without changing the baseline arrangement.

\textbf{Expected behavior.} Distinguish the baseline authority question from voluntary one-off adjustments, compare signatures and provenance, and recommend written verification or professional clarification rather than declaring the app or printed plan automatically controlling.

\textbf{Failure contrast.} ``Follow the flexible app plan because it is the shared platform'' ignores the authentication and error signals.

\textbf{Case S3: \texttt{p07\_service\_status} --- written evidence versus repeated verbal claims.} \textbf{Why selected.} This is a compact source-count trap: two bank representatives agree verbally, but a printed classification summary points the other way. It demonstrates why source majority is not a valid reliability rule and gives a clear, practical written-verification path.

\textbf{Query.} ``Help me get clear confirmation of whether my main bank account is set up as a business account so I can handle store transactions and family support money properly.''

\textbf{Memory bank.}
\begin{itemize}[leftmargin=1.35em,itemsep=1pt,topsep=2pt]
\item \emph{Accountant claim:} mixed store and household use does not qualify the account for the small-business tier.
\item \emph{Branch claim:} a branch representative says the account already has business-tier benefits.
\item \emph{Plausibility context:} frequent movement between store revenue and family expenses makes either classification superficially plausible.
\item \emph{Documentary evidence:} the accountant provides a printed internal summary classifying the account as personal and affecting loan eligibility.
\item \emph{Second verbal claim:} a phone agent also reports a business designation and business-tier deposit rates.
\item \emph{Change-risk signal:} a neighboring store owner experienced an unannounced status change at the same bank.
\item \emph{Access constraint:} Amina has only limited early-morning time for a formal follow-up.
\end{itemize}
\textbf{Distractors.} Extended branch hours, transaction alerts, prior identity-verified servicing, phone-support logs, e-statements, and appointment booking help obtain follow-up but do not establish the tier classification.

\textbf{Pattern.} \emph{Printed personal-tier classification versus two agreeing verbal business-tier assurances.} The count of verbal sources does not supersede documentary evidence, especially when status may change silently.

\textbf{Unresolved slot.} The bank's current formal classification and the associated business-tier and loan eligibility.

\textbf{Expected behavior.} Preserve both claims, request a formal written classification statement, and use Amina's feasible follow-up window rather than treating either verbal account as final.

\textbf{Failure contrast.} ``It must be business tier because two representatives confirmed it'' mistakes repetition for authoritative confirmation.

\textbf{Case S4: \texttt{p40\_event\_scheduling} --- conflict-blind synthesis.} \textbf{Why selected.} This retained stress case exposes a different SCC failure: models may merge mutually exclusive schedules into a polished but impossible weekly plan. The coordinator report and sister's observation support one schedule, whereas preference and work-planning memories support another; a safe response must first verify the actual schedule.

\textbf{Query.} ``Can you help me set up something that works for scheduling my dialysis, clinical trial visits, and bookkeeping appointments so my week feels manageable?''

\textbf{Memory bank.}
\begin{itemize}[leftmargin=1.35em,itemsep=1pt,topsep=2pt]
\item \emph{Preferred schedule:} Selene says Tuesday/Thursday mornings would preserve weekends and recovery time.
\item \emph{Coordinator schedule:} a research coordinator reports Monday/Wednesday/Friday afternoons as the actual dialysis routine for six months.
\item \emph{General clinic claim:} Selene also says clinic appointments occupy mornings.
\item \emph{Independent observation:} her sister observes exhaustion after Monday/Wednesday/Friday evening returns.
\item \emph{Work-planning contradiction:} Selene schedules Monday/Wednesday client meetings as if those were appointment-free days.
\item \emph{Post-treatment preference:} work from home is preferred after treatment but does not resolve which days treatment occurs.
\end{itemize}
\textbf{Distractors.} Monthly nephrology planning, trial calls, grocery timing, bookkeeping document deadlines, trial lab draws, and post-clinic rest are all schedule-relevant but cannot verify the disputed dialysis days.

\textbf{Pattern.} \emph{Tuesday/Thursday-morning preference and planning versus coordinator-reported, sister-corroborated Monday/Wednesday/Friday-afternoon routine.} The conflict cannot be safely solved by combining all blocks into one calendar.

\textbf{Unresolved slot.} Selene's current actual dialysis schedule and whether the coordinator record remains current.

\textbf{Expected behavior.} Identify the incompatible schedules, refrain from composing a definitive combined calendar, and request confirmation from Selene or the clinic/coordinator before assigning trials and bookkeeping around treatment.

\textbf{Failure contrast.} A plan that schedules Tuesday/Thursday dialysis while also preserving Monday/Wednesday/Friday post-treatment recovery merges contradictory facts into an infeasible schedule.

\subsection{Pipeline Observability Contrasts}
\label{app:pipeline_observability}

The Oracle cards above show complete canonical banks. Pipeline evaluation instead asks whether a retrieved bank preserves the target conflict. \textsc{Full} means that all conflict-bearing relations are visible; \textsc{partial} means that some target-specific tension survives but at least one decisive relation is missing; and \textsc{none} means that the target conflict is not observable, even if the retrieved memories remain useful and faithful for answering the query. Table~\ref{tab:pipeline_observability_examples} gives compact contrasts from the pipeline audit. The States column lists the Stage-1 labels for Mem0 / Letta / A-mem / MemOS, in that order.

{\footnotesize
\begin{xltabular}{\linewidth}{@{}l c l Y Y@{}}
\caption{Illustrative Pipeline observability contrasts. \textsc{None} does not mean that a retrieved bank is useless; it means that the benchmark target conflict is absent, so D1 and D2 are not applicable.}\label{tab:pipeline_observability_examples}\\
\toprule
\textbf{Case} & \textbf{Axis} & \textbf{States} & \textbf{What is preserved or lost} & \textbf{Interpretation} \\
\midrule
\endfirsthead
\multicolumn{5}{c}{\tablename~\thetable\ (continued)}\\
\toprule
\textbf{Case} & \textbf{Axis} & \textbf{States} & \textbf{What is preserved or lost} & \textbf{Interpretation} \\
\midrule
\endhead
\midrule
\multicolumn{5}{r}{\footnotesize\emph{continued on next page}}\\
\endfoot
\bottomrule
\endlastfoot
\texttt{p09} & CPC & P/F/P/N & A post-shift pause can support useful, grounded advice, but the competence or next-day-obligation partition may be absent. & A faithful and actionable response can coexist with NONE observability; D1/D2 are then not applicable rather than zero. \\
\texttt{p24} & BOC & N/F/P/P & The final preference for solo, untracked walking does not expose the earlier shared-app and fridge-log phases. & FULL requires the multi-phase structure--surveillance--withdrawal trajectory, not merely the current preference. \\
\texttt{p01} & BOC & N/P/P/P & Scheduling and note-taking details are relevant while later recurrence remains incompletely represented. & PARTIAL can preserve meaningful tension without enough episodes for a complete trajectory diagnosis. \\
\texttt{p40} & SCC & P/F/P/N & General paperwork confusion is not the attributed liaison-versus-attorney disagreement about authorization. & FULL requires competing claims, their sources, and enough reliability context for source-conflict reasoning. \\
\end{xltabular}
}

\noindent\textit{Abbreviations:} F = \textsc{Full}; P = \textsc{Partial}; N = \textsc{None}. State order: Mem0 / Letta / A-mem / MemOS.

\end{document}